\pdfoutput=1

\documentclass[11pt]{article}

\newif\ifarxiv
\arxivtrue

\ifarxiv
    \usepackage[final]{acl}
\else
    \usepackage[review]{acl}
\fi

\usepackage{times}
\usepackage{latexsym}
\usepackage{amsmath,amssymb,amsthm}
\usepackage{bbm}
\usepackage{mathabx,mathrsfs}
\usepackage{relsize}
\PassOptionsToPackage{hyphens}{url}\usepackage{hyperref}
\usepackage{xurl}
\usepackage[T1]{fontenc}

\usepackage{CJKutf8}
\usepackage[utf8]{inputenc}

\usepackage{microtype}

\usepackage{inconsolata}

\usepackage{natbib}

\usepackage{todonotes}
\usepackage{microtype}
\usepackage{pifont}
\usepackage{bm}
\usepackage{amsmath}
\usepackage{amssymb}
\usepackage{autobreak}
\usepackage{float}
\usepackage{subcaption}
\usepackage{paralist}
\usepackage{xpinyin}
\usepackage{scalerel}
\usepackage{tikz}
\usepackage{tikz-dependency}
\usepackage{tikz-qtree}
\usepackage{tikz-layers}
\usepackage{tikzsymbols}
\usepackage{pgfplots}
\usepackage{graphicx}
\usepackage{adjustbox}
\usepgflibrary{arrows.meta}
\usepgfplotslibrary{fillbetween, statistics, groupplots}
\usetikzlibrary{fit, calc, plotmarks, decorations.pathreplacing, decorations.markings, positioning, arrows.meta, shapes.geometric, backgrounds, graphs, intersections}
\pgfplotsset{compat=1.18}
\usepackage{siunitx}
\usepackage{rotating}
\usepackage{booktabs}
\usepackage{tcolorbox}
\tcbuselibrary{raster}
\usepackage{xcolor,colortbl}
\usepackage{nicematrix}
\usepackage{mleftright}
\usepackage{multirow}
\usepackage{cleveref}

\newcommand\Letter{{\fontfamily{mvs}\fontencoding{U}\selectfont\char66}}

\def\myemoji@insert#1{\scalerel*{\includegraphics[page=#1]{assets/myemoji-assets.pdf}}{X}}
\newcommand\orange{1}
\newcommand\coconut{2}
\def\hwemoji@insert#1{\scalerel*{\includegraphics[page=#1]{assets/hwemoji-assets.pdf}}{X}}
\newcommand\mailemoji{\hwemoji@insert{3610}}
\newcommand\newemoji{\hwemoji@insert{12}}

\newcommand\snowflakeemoji{\hwemoji@insert{3644}}

\definecolor{thinking_yellow}{RGB}{187,122,44}
\definecolor{json_blue}{RGB}{15, 89, 164}
\definecolor{json_red}{RGB}{192, 44, 53}
\definecolor{figure_green}{RGB}{32, 137, 77}
\definecolor{figure_blue}{RGB}{52, 108, 156}
\definecolor{figure_red}{RGB}{192, 44, 53}
\definecolor{figure_orange}{RGB}{250, 126, 35}
\definecolor{table_blue}{RGB}{92, 179, 204}
\definecolor{table_red}{RGB}{238, 63, 77}
\definecolor{table_orange}{RGB}{250, 126, 35}
\definecolor{figure_light_green}{RGB}{198, 223, 200}
\definecolor{figure_light_blue}{RGB}{208, 223, 230}
\definecolor{figure_light_red}{RGB}{192, 44, 53}
\definecolor{figure_light_gray}{RGB}{220, 220, 220}
\definecolor{figure_gray}{RGB}{160, 160, 160}

\newcommand\jsonkey[1]{\texttt{\textbf{\textcolor{json_blue}{#1}}}}

\newcommand\substring[1]{\ensuremath{{\boldsymbol{#1}}}}

\newcommand\return{\texttt{\textcolor{gray}{\textbackslash n}}}

\newcommand\wz{\phantom{0}}
\newcommand\wc{\phantom{,}}

\newcommand\correct[1]{\textcolor{figure_blue}{{#1}}}
\newcommand\wrong[1]{\textcolor{figure_red}{{{#1}}}}

\newtcolorbox{promptbox}[3]{%
    left=4pt,
    right=4pt,
    top=4pt,
    bottom=4pt,
    boxsep=3pt,
    colback={#3},
    colframe={#2},
    title={#1},
}

\usepackage{contour}
\usepackage[normalem]{ulem}
\usepackage{soul}

\setuldepth{Berlin}%

\NiceMatrixOptions
       {
         custom-line =
           {
             letter = ; ,
             command = hdashedline ,
             ccommand = cdashedline ,
             tikz = dashed
    } 
}

\pgfdeclarelayer{background}  %
\pgfdeclarelayer{above}       %
\pgfsetlayers{background,main,above}  %

\newcommand \footnotetextonly[1]
{
    \let \backupfootnote \thefootnote
    \let \thefootnote \relax
    \footnotetext{#1}
    \let \thefootnote \backupfootnote
    \let \backupfootnote \imreallyundefinedcommand
}

\title{
    A Training-free LLM-based Approach\\to General Chinese Character Error Correction
}

\ifarxiv
    \author{Houquan Zhou\rlap{$^{\orange}$},\ \ Bo Zhang\rlap{$^{\coconut}$},\ \ Zhenghua Li\rlap{$^{\orange\text{\Letter}}$} \\
    {\bf Ming Yan\rlap{$^{\coconut}$},\ \  Min Zhang$^{\orange}$}  \\%
    $^{\orange}$School of Computer Science and Technology, 
    Soochow University, China \\%
    \texttt{hqzhou@stu.suda.edu.cn}, \texttt{\{zhli13,minzhang\}@suda.edu.cn}\\%
    $^{\coconut}$Alibaba Group, China\\%
    \texttt{\{klayzhang.zb,ym119608\}@alibaba-inc.com}}
\else
    \author{Anonymous}
\fi

\pgfplotsset{compat=1.17}
\begin{document}
\begin{CJK}{UTF8}{gkai}
    \maketitle
    \ifarxiv%
        \footnotetextonly{\!\!\Letter\ Zhenghua Li is the corresponding author.}
    \fi%

    \begin{abstract}
    Chinese spelling correction (CSC) is a crucial task that aims to correct character errors in Chinese text.
    While conventional CSC focuses on character substitution errors caused by mistyping, two other common types of character errors, missing and redundant characters, have received less attention.
    These errors are often excluded from CSC datasets during the annotation process or ignored during evaluation, even when they have been annotated.
    This issue limits the practicality of the CSC task.
    To address this issue, we introduce the task of General Chinese Character Error Correction (C2EC), which focuses on all three types of character errors.
    We construct a high-quality C2EC benchmark by combining and manually verifying data from CCTC and Lemon datasets.
    We extend the training-free prompt-free CSC method to C2EC by using Levenshtein distance for handling length changes and leveraging an additional prompt-based large language model (LLM) to improve performance.
    Experiments show that our method enables a 14B-parameter LLM to be on par with models nearly 50 times larger on both conventional CSC and C2EC tasks, without any fine-tuning\rlap{.}%
    \ifarxiv%
    \else%
        \footnote{Our anonymized code is available at \url{https://anonymous.4open.science/r/simple-c2ec}.}
    \fi
\end{abstract}

    \section{Introduction}
Given an input sentence $\substring{x} = x_1, x_2 \cdots x_n$, the task of conventional Chinese spelling correction (CSC) aims to produce a new sentence of the same length, denoted as $\substring{y} = y_1, y_2 \cdots y_n$, where each misspelled character (\textit{e.g.,} $x_i$) is replaced with a correct one (\textit{e.g.,} $y_i$).
Spelling errors in text can cause misunderstandings, damage authenticity, and even lead to unnecessary financial losses, making automatic correction a crucial task in Chinese Natural Language Processing (NLP) \cite{wu-etal-2023-rethinking,zhou-etal-2024-simple,li-etal-2024-cllm,dong-etal-2024-rich,liu-etal-2024-arm}.

Spelling errors mainly arise from two sources:
The first source is typing mistakes.
Chinese characters often have multiple visually or phonetically similar variants, making it easy to select an incorrect one when using input methods \cite{hu-etal-2024-cscd}.
The second source is automatic text conversion errors.
When using automatic speech recognition (ASR) or optical character recognition (OCR), systems may also introduce incorrect characters during the conversion process \cite{wang-etal-2018-hybrid}.

\begin{figure}[tb!]
    \newtcolorbox{scopebox}[3]{%
        left=0pt,
        right=0pt,
        top=0pt,
        bottom=0pt,
        boxsep=3pt,
        colback={#3},
        colframe={#2},
        title={#1},
    }
    \centering
    \scriptsize
    \begin{scopebox}{General Chinese Character Error Correction}{black}{black!2}
        \begin{scopebox}{Misspelling Character (\textit{Conventional Chinese Spelling Correction})}{figure_blue!70}{white}
            我这学期选修了[\wrong{羽矛球}$\rightarrow$\correct{羽毛球}]课。

            \textit{I have taken [\wrong{bad\textbf{n}inton}$\rightarrow$\correct{bad\textbf{m}inton}] as an elective this semester.}
        \end{scopebox}
        \begin{scopebox}{Missing Character (\textit{New})}{figure_green!85}{white}
            我这学期选修了[\wrong{羽球}$\rightarrow$\correct{羽毛球}]课。

            \textit{I have taken [\wrong{badinton}$\rightarrow$\correct{bad\textbf{m}inton}] as an elective this semester.}
        \end{scopebox}
        \begin{scopebox}{Redundant Character (\textit{New})}{figure_green!85}{white}
            我这学期选修了[\wrong{羽矛毛球}$\rightarrow$\correct{羽毛球}]课。

            \textit{I have taken [\wrong{badmin\textbf{t}ton}$\rightarrow$\correct{badminton}] as an elective this semester.}
        \end{scopebox}
    \end{scopebox}

    \caption{
        Scope of the general Chinese character error correction.
        The correct sentence should be “我这学期选修了羽毛球课。” (\textit{I have taken badminton as an elective this semester.}).
    }
    \label{fig:research_scope}
\end{figure}

Beyond spelling errors, character errors involving missing and redundant characters are also common \cite{he-etal-2023-umrspell}.
These errors can be caused by the same reasons as spelling errors.
For instance, when typing ``曲安'' (\textit{qū ān}) in ``醋酸曲安奈德'' (\textit{Triamcinolone Acetonide}), the input method might consider ``\textit{qū ān}'' as a single character ``圈'' (\textit{quān}), resulting in a misspelled and missing character error.\footnote{A real case from the Lemon \textit{Mec} subset.}
Additionally, repetitive sentence editing, such as when revising messages, social media posts, or emails, can easily introduce missing and redundant errors.
Consider this example: ``我参加的项目是\wrong{羽球}。'' (\textit{The project I participated in is \wrong{badmton}}), where ``毛'' was omitted from ``羽毛球'' (\textit{badminton}).
The missing ``毛'' likely occurred during editing: when correcting a mistyped ``求'' to ``球'', the user accidentally deleted two characters before retyping ``球''.
Such errors often go unnoticed without thorough proofreading.
For clarity, we refer to misspelling, missing, and redundant errors as \textbf{General Chinese Character Errors} (C2E).

Recent Chinese text correction competitions, CTC 2021 \cite{zhao-etal-2022-overview}, Midu-CTC\footnote{\url{https://aistudio.baidu.com/competition/detail/404/0/introduction}}, and Kingsoft-CTC\footnote{\url{https://datastudio.wps.cn/matchcenter/competition/1/introduction}}, have designed error distributions to reflect real-world scenarios.
In these competitions, C2E constitute 79.4\% to 87.3\% of errors (Table~\ref{tab:data_statistics_main}).
This indicates that C2E are more common than complex errors like grammatical mistakes, logical inconsistencies, or ambiguity.
Compared to conventional CSC, addressing C2E is more practical as it covers a broader range of common errors.

Thus, we believe that \textbf{General Chinese Character Error Correction} (C2EC) deserves more attention from the Chinese text correction community.
While \citet{he-etal-2023-umrspell} previously studied C2EC, they created synthetic ECMR-2023 by placing random errors into correct sentences.
There is a lack of a dataset specifically focusing on C2E with real-world errors.
To fill this gap, we build a new C2EC dataset using two existing datasets containing real-world errors: CCTC \cite{wang-etal-2022-cctc} and Lemon \cite{wu-etal-2023-rethinking}.
We carefully verified the data to ensure data quality and annotation consistency, resulting in 1,995 sentences for development and 5,711 for testing.

Recent work has shown the power of large language models (LLMs) for CSC \cite{dong-etal-2024-rich,li-etal-2024-cllm,zhou-etal-2024-simple}.
Notably, by combining an LLM, which evaluates fluency, with a Hamming distance to prevent over-correction, the training-free prompt-free framework (\texttt{TfPf}) \cite{zhou-etal-2024-simple} achieves strong results without any training.
We extend this approach to C2EC by using Levenshtein distance to handle missing and redundant errors, and incorporating prompt-based probability scoring for better performance.

Experiments show that our approach achieves large improvements over the \texttt{TfPf} baseline on both conventional CSC and C2EC datasets, and even outperforms the supervised fine-tuned counterparts on various domains.
It is also worth mentioning that our approach enables a 14B parameter model to achieve competitive performance with models nearly 50 times larger without any training.

\ifarxiv%
    Our code is available at \url{https://github.com/Jacob-Zhou/simple-csc/tree/v2.0.0}.
\fi%

    \section{The C2EC Task}
\label{sec:gcsc_benchmark}
\begin{table*}[tb!]
    \centering
    \setlength{\tabcolsep}{3.2pt}
    \scalebox{0.9}{
        \begin{NiceTabular}{lccc;cccc;c}
            \toprule
            \Block[l]{2-1}{\textbf{Datasets}}       & \Block[c]{2-1}{\textbf{\# Sent}} & \Block[c]{2-1}{\textbf{\% Err.}                                                                                                                                                                  \\\textbf{Sent}} & \Block[c]{2-1}{\textbf{Avg.}\\\textbf{Len.}} & \Block[c]{1-4}{\textbf{\# Char. Error}} &              &              &              & \Block[c]{2-1}{\textbf{\# Other}\\\textbf{Error}} \\
                                                    &                                  &                                 &              & \textbf{SUB}                   & \textbf{RED}            & \textbf{MIS}           & \textit{All}                       &                        \\
            \midrule
            \textbf{CTC21} {\textit{Qua}}           & \wc\wz\wz969                     & \wz49.5                         & 48.9         & \wc\wz\wz280   (\textit{65.6}) & \wz59 (\textit{13.8})   & \wz88  (\textit{20.6}) & \wz\wz427   [\textit{79.4}]        & 111 [\textit{20.6}]    \\
            \textbf{MiduCTC} {\textit{Val}}         & \wz1,014                         & \wz50.7                         & 57.3         & \wc\wz\wz389   (\textit{82.2}) & \wz35 \wz(\textit{7.4}) & \wz49  (\textit{10.4}) & \wz\wz473   [\textit{85.7}]        & \wz79  [\textit{14.3}] \\
            \textbf{KingsoftCTC} {\textit{Labeled}} & \wz2,000                         & 100.0                           & 79.4         & \wz2,538  (\textit{75.1})      & 607 (\textit{18.0})     & 236 \wz(\textit{7.0})  & \wz3,381  [\textit{87.3}]          & 493 [\textit{12.7}]    \\
            \midrule
            \textbf{CCTC} {\textit{Test}}           & 13,037                           & \wz\wz9.3                       & 41.1         & \wc\wz\wz933   (\textit{74.9}) & 146 (\textit{11.7})     & 166 (\textit{13.3})    & \wz1,245  [\textit{92.6}]          & 100 \wz[\textit{7.4}]  \\
            \textbf{Lemon} {\textit{All}}           & 22,252                           & \wz50.0                         & 35.4         & 12,094 (\textit{94.8})         & 369 \wz(\textit{2.9})   & 288 \wz(\textit{2.3})  & 12,751   \phantom{[\textit{00.0}]} & --                     \\
            \midrule
            \textbf{ECMR-2023} {\textit{Test}}      & \wz1,000                         & \wz\texttt{N/A}                 & \texttt{N/A} & \wz3,105  (\textit{81.6})      & 456 (\textit{12.0})     & 246 \wz(\textit{6.5})  & \wz3,807 \phantom{[\textit{00.0}]} & --                     \\
            \hdashedline
            \textbf{C2EC} {\textit{Dev}}            & \wz1,995                         & \wz50.1                         & 51.9         & \wc\wz\wz818   (\textit{74.5}) & 148 (\textit{13.5})     & 132 (\textit{12.0})    & \wz1,098 \phantom{[\textit{00.0}]} & --                     \\
            \textbf{C2EC} {\textit{Test}}           & \wz5,711                         & \wz49.9                         & 41.9         & \wz2,397  (\textit{72.6})      & 462 (\textit{14.0})     & 442 (\textit{13.4})    & \wz3,301 \phantom{[\textit{00.0}]} & --                     \\
            \bottomrule
        \end{NiceTabular}
    }
    \caption{
    The statistics of the datasets. Numbers in the parentheses ``()'' are the percentages of the total number of character errors. Numbers in the brackets ``[]'' are the percentages of the total number of all errors.
    }
    \label{tab:data_statistics_main}
\end{table*}

\subsection{Task Definition}
Given an input sentence $\substring{x} = x_1,x_2\cdots{}x_n$, the task of C2EC aims to correct character errors and produce a corrected sentence $\substring{y} = y_1,y_2\cdots{}y_m$. Unlike conventional CSC, where input and output lengths must match, C2EC allows $\substring{y}$ to have a different length from $\substring{x}$. In addition to character substitutions, C2EC addresses two additional error types:
\begin{inparaenum}[\itshape a)]
    \item \textbf{Missing Character}: where characters are missing from the input, e.g., ``羽球'' missing ``毛'' from ``羽毛球'' (\textit{badminton}).
    \item \textbf{Redundant Character}: where unexpected extra characters appear in the input, e.g., the extra ``矛'' in ``羽\wrong{矛}毛球''.
\end{inparaenum}

\subsection{Construction of C2EC Dataset}
Rather than using rule-based synthesis like \citet{he-etal-2023-umrspell}, we construct our dataset from existing datasets to better reflect real-world error patterns.

\paragraph{Data Selection and Division}
We build the C2EC dataset from two high-quality sources:
\begin{inparaenum}[\itshape a)]
    \item \textbf{CCTC} \cite{wang-etal-2022-cctc}: A comprehensive dataset of \textasciitilde{}25,000 sentences covering both character errors and complex errors from diverse sources. While this dataset includes C2E errors, it was not specifically designed to focus on C2EC.
    \item \textbf{Lemon} \cite{wu-etal-2023-rethinking}: A CSC dataset containing \textasciitilde{}500 sentences with missing and redundant character errors. Although it contains sentences with C2E errors, previous works excluded these when evaluating CSC methods \cite{wu-etal-2023-rethinking,liu-etal-2023-chinese,zhou-etal-2024-simple,li-etal-2024-cllm}.
\end{inparaenum}
We use CCTC's training set for development and combine the test sets from both CCTC and Lemon for testing.

\paragraph{Data Resampling}
To achieve a more balanced and natural error distribution, we perform two resampling steps:
\begin{inparaenum}[\itshape a)]
    \item We balance the ratio of correct to incorrect sentences to 1:1, reducing the original 91\% correct sentence bias in CCTC.
    \item We adjust Lemon's error type distribution to match CCTC's (\textasciitilde{}75\% substitution, \textasciitilde{}25\% missing/redundant), addressing the bias of Lemon toward substitution errors (95\%).
\end{inparaenum}
Throughout resampling, we maintain balanced distributions across domains and correct-incorrect sentence pairs.

\paragraph{Quality Control}
To maintain focus on character-level errors and keep the dataset clean, we apply two data cleaning processes:
\begin{inparaenum}[\itshape a)]
    \item Automatically remove sentences with complex errors, specifically those errors that involve continuous insertions or deletions (this removed 91 and 90 sentences from the development and test sets, respectively).
    \item Manually verify sentences by native Chinese speakers to remove sentences that:
    \begin{inparaenum}[\itshape i.]
        \item Have incorrect annotations;
        \item Have complex grammatical errors;
        \item Have multiple reasonable corrections;
        \item Are ambiguous or difficult to understand.
    \end{inparaenum}
\end{inparaenum}
Four native Chinese speakers familiar with the CSC task performed the manual verification.
Each sentence was reviewed by at least two annotators, with a third resolving any disagreements.
The inter-annotator agreement reached 97.08\%.
This process removed 111 test set and 31 development set sentences.\footnote{Several examples of discarded sentences are provided in the appendix~\ref{sec:discard_cases}.}

\subsection{Data Statistics}
The final dataset, as shown in Table~\ref{tab:data_statistics_main}, contains 1,995 development and 5,711 test sentences, with approximately half being error-free. The development set contains 1,098 errors (\textasciitilde{}1.1 per erroneous sentence), while the test set has 3,301 errors (\textasciitilde{}1.2 per erroneous sentence).
In the test set, 72.6\% of errors are misspellings, 14.0\% are redundant characters, and 13.4\% are missing characters.

    \section{The Baseline TfPf Approach}
\label{sec:baseline_tfpf_approach}
The training-free prompt-free framework (\texttt{TfPf}) \cite{zhou-etal-2024-simple} combines a large language model with a distance metric:
\begin{equation}
    \begin{aligned}
        s(\substring{x}, \substring{y}) = \log p_{\mathtt{LLM}}(\substring{y}) - \mathtt{Dist}(\substring{x}, \substring{y})
        \label{eq:tfpf_score}
    \end{aligned}
\end{equation}
The first term $p_{\mathtt{LLM}}(\substring{y})$ represents a large language model that models the probability of the correct sentence $\substring{y}$, ensuring the correction is fluent from a language perspective.
The second term $\mathtt{Dist}(\substring{x}, \substring{y})$ measures the Hamming distance\footnote{The original \texttt{TfPf} paper refers to this part as the distortion model. However, from a mathematical perspective, the distortion model is equivalent to minus the weighted Hamming distance between $\substring{x}$ and $\substring{y}$. Details are in Appendix~\ref{subsec:distortion_model_of_tfpf_as_a_weighted_hamming_distance}.} between $\substring{x}$ and $\substring{y}$, preventing the model from making too many changes or inserting/deleting characters in the input sentence to achieve fluency.

\section{Our Training-free Approach for C2EC}
\label{sec:our_approach}

We extend the \texttt{TfPf} framework with the following equation:
\begin{equation}
    \begin{aligned}
        s(\substring{x}, \substring{y}) =\  & {\log p_{\mathtt{LLM}}(\substring{y}\mid \mathtt{Prompt}(\substring{x}))}                         \\
                                            & + \log p_{\mathtt{LLM}}(\substring{y}) - \mathtt{Dist}_{\mathtt{L}}(\substring{x}, \substring{y})
        \label{eq:gcsc_score}
    \end{aligned}
\end{equation}
where $\mathtt{Prompt}(\cdot)$ is the prompt template.

Our approach makes three key extensions to the \texttt{TfPf} framework:
\begin{inparaenum}[\itshape a)]
    \item We allow the input sentence $\substring{x}$ to have a different length from the output sentence $\substring{y}$.
    \item We replace the original Hamming distance metric with Levenshtein distance to handle varying input and output lengths. We use $\mathtt{Dist}_{\mathtt{L}}$ to distinguish it from the original distance metric $\mathtt{Dist}$.
    \item We incorporate an additional prompt-based probability $p_{\mathtt{LLM}}(\substring{y}\mid \mathtt{Prompt}(\substring{x}))$ in the score function to improve performance.
\end{inparaenum}

\subsection{Incremental Weighted Levenshtein Distance for Generation}
Following \texttt{TfPf}, we use a weighted distance metric to measure the distance between $\substring{x}$ and $\substring{y}$.
We classify each edit operation into different types and assign a specific weight to each type.
Table~\ref{tab:type_example} shows the weights used in our work.
For operations already defined in \texttt{TfPf}, we adopt their original weights\footnote{\url{https://github.com/Jacob-Zhou/simple-csc/blob/main/configs/default_config.yaml}}.
For the two new operation types, \texttt{Insert} and \texttt{Delete}, we set weights to 8.5 and 9.0, respectively, after grid search on development sets.

Large language models generate output incrementally, while Levenshtein distance requires solving a dynamic programming problem over the entire sentence.

To mitigate this incompatibility, we introduce an incremental Levenshtein distance.
For a partial output candidate $\substring{y}_{\le j}$, we define the partial distance as the Levenshtein distance between $\substring{y}_{\le j}$ and a corresponding prefix of the input sentence $\substring{x}_{\le b}$.
The incremental Levenshtein distance after generating a new character $y_j$ is:
\begin{equation}
    \begin{aligned}
         & \Delta_{\mathtt{Dist}_{\mathtt{L}}}(\substring{x}, a, b, \substring{y}_{<j}, y_j)                                                                               \\
         & \quad = \mathtt{Dist}_{\mathtt{L}}(\substring{x}_{\le b}, \substring{y}_{<j} \circ y_j) - \mathtt{Dist}_{\mathtt{L}}(\substring{x}_{\le a}, \substring{y}_{<j})
    \end{aligned}
\end{equation}
where $\circ$ denotes concatenation, and $a$ and $b$ are the end indices of the input prefix before and after generating $y_j$, respectively.
\begin{table}[tp!]
    \setlength{\tabcolsep}{4.5pt}
    \centering
    \begin{tabular}{lcc}
        \toprule
        \textbf{Edit Type}                  & \textbf{Example}                                 & \textbf{Weight} \\
        \midrule
        \texttt{Keep}                       & 机器 (\textit{jī qì})                              & \wz0.04         \\
        \midrule
        \texttt{Substitute}                                                                                      \\
        \quad\texttt{Same \rlap{Pinyin}}    & 基器  (\textit{jī qì})                             & \wz3.75         \\
        \quad\texttt{Similar \rlap{Pinyin}} & 七器  (\textit{\textcolor{figure_blue}{q}ī qì})    & \wz4.85         \\
        \quad\texttt{Similar \rlap{Shape}}  & \textcolor{figure_blue}{仉}器  (\textit{zhǎng qì}) & \wz5.40         \\
        \quad\texttt{Other Similar}         & 金器  (\textit{jī\textcolor{figure_blue}{n} qì})   & \wz8.91         \\
        \midrule
        \texttt{Insert}                     & 机器 $\rightarrow$ 机器\textcolor{figure_blue}{人}    & \wz8.50         \\
        \texttt{Delete}                     & \textcolor{figure_blue}{机}器 $\rightarrow$ 器      & \wz9.00         \\
        \bottomrule
    \end{tabular}
    \caption{
        Examples of the different edit types of the corrected token ``机器'' (\textit{jī}).
        We adopt the weight of substitution from \citet{zhou-etal-2024-simple}.
    }
    \label{tab:type_example}
\end{table}

\subsection{Extra Use of Prompt-based LLM}
\paragraph{A Reinforcement Learning Perspective}
The use of two LLMs in Equation~\ref{eq:gcsc_score} can be understood from a reinforcement learning perspective.
In this view, the \texttt{TfPf} framework serves as a reward function that guides the generation process.
While the prompt-based LLM acts as a reference model that produces correction probabilities based on the prompt, the pure LLM evaluates the fluency of generated sequences.
Details on this perspective are provided in Appendix~\ref{sec:reinforcement_learning_perspective}.

\paragraph{Selection of LLM}
\label{sec:selection_of_llm}
To save GPU memory, we use the same model for both the prompt-based LLM and language model functions.
When GPU memory is sufficient, different models could be used for each function.

We use the base version of the LLM for both the prompt-based LLM and the language model, as our experiments show that it is more robust than instruction-tuned versions across different prompt templates and achieves better overall performance.

\paragraph{Prompt Template}
\label{sec:prompt_template}

\begin{figure}[tb!]
    \centering%
    \begin{promptbox}{\texttt{Minimal Prompt}}{black}{white}
        \footnotesize{
            纠正下面句子中的错别字以及多字少字错误，并给出修改后的句子。\return

            输入：\jsonkey{\{INPUT: $\boldsymbol{x}$\}}\return

            输出：
        }
        \tcblower
        \footnotesize
        Correct the typos and redundant or missing characters in the following sentence, and provide the sentence after correction.\return

        \footnotesize
        Input: \jsonkey{\{INPUT: $\boldsymbol{x}$\}}\return

        \footnotesize
        Output:

    \end{promptbox}
    \caption{
        A minimal prompt template used in our method.
    }
    \label{fig:minimal_prompt}
\end{figure}

We designed two prompt templates: a minimal prompt (\texttt{Minimal}) based on \citet{li-etal-2024-cllm} and a more sophisticated prompt (\texttt{Detailed}) based on \citet{li-etal-2023-ineffectiveness,zhou-etal-2024-simple}.
Figure~\ref{fig:minimal_prompt} shows the minimal prompt, which contains only essential task description and input sentence.
The detailed prompt, shown in Figure~\ref{fig:detail_prompt} in Appendix~\ref{sec:prompt_template}, additionally includes task-specific system prompts, format requirements, and notes.

Our experiments show that the base version of LLM performs better with the minimal prompt, while instruction-tuned LLM works better with the detailed prompt.
Based on overall performance, we adopt the minimal prompt in this work.

\subsection{Token-based Generation and Beam Search}
Large language models use multi-character tokenization for Chinese.
A token $\substring{t}_k$ may contain multiple characters.
For clarity, we define a partial output candidate at generation step $k$ as $\substring{y}_{\le j}=y_1\cdots{}y_j=\substring{t}_1\cdots\substring{t}_k$.

At each generation step $k$, we calculate the token score $\substring{t}_k$ as:
\begin{equation*}
    \begin{aligned}
         & s(\substring{x}, \substring{t}_1\dots\substring{t}_k, b) = s(\substring{x}, \substring{t}_1\dots\substring{t}_{k-1}, a)          \\
         & \qquad + \log p_{\mathtt{LLM}}(\substring{t}_k\mid \mathtt{Prompt}(\substring{x}) \circ \substring{t}_1\dots\substring{t}_{k-1}) \\
         & \qquad + \log p_{\mathtt{LLM}}(\substring{t}_k\mid \substring{t}_1\dots\substring{t}_{k-1})                                      \\
         & \qquad - \Delta_{\mathtt{Dist}_{\mathtt{L}}}(\substring{x}, a, b, \substring{t}_1\dots\substring{t}_{k-1}, \substring{t}_k)
    \end{aligned}
\end{equation*}
where $\Delta_{\mathtt{Dist}_{\mathtt{L}}}(\substring{x}, a, b, \substring{t}_1\dots\substring{t}_{k-1}, \substring{t}_k)$ is the incremental Levenshtein distance after generating multiple characters.

We use beam search to find the final output.
That is, at each generation step $k$, we expand the candidates with all possible tokens $\substring{t}_k$ and end indices $b$ and only keep the top $K$ candidates with the highest scores.
Following \texttt{TfPf}, the beam size is set to~8.

\subsection{The Full Model}
Following \texttt{TfPf}, we also adopt their length and faithfulness rewards in the score function:
\begin{equation}
    \begin{aligned}
         & s(\substring{x} ; \substring{t}_1\dots\substring{t}_k; b) = s(\substring{x}, \substring{t}_1\dots\substring{t}_{k-1}, a)                    \\
         & \qquad + \log p_{\mathtt{LLM}}(\substring{t}_k\mid \mathtt{Prompt}(\substring{x}) \circ \substring{t}_1\dots\substring{t}_{k-1})            \\
         & \qquad + \log p_{\mathtt{LLM}}(\substring{t}_k\mid \substring{t}_1\dots\substring{t}_{k-1})                                                 \\
         & \qquad + \lambda \left(\begin{array}{c}
                                          - \Delta_{\mathtt{Dist}_{\mathtt{L}}}(\substring{x}, a, b, \substring{t}_1\dots\substring{t}_{k-1}, \substring{t}_k) \\
                                          +                                                                                                                    \\
                                          \alpha\left( \mathtt{len}(\substring{t}_k)-1 \right)
                                      \end{array}\right)
    \end{aligned}
    \label{eq:final_score}
\end{equation}
where $\lambda = (1 + H_{\mathtt{LLM}}(\cdot))$ is the faithfulness reward that dynamically increases the distance impact when the language model is uncertain, and $\alpha(\mathtt{len}(\substring{t}_k)-1)$ is the length reward encouraging longer token selection.

    \section{Experimental Setup}
\subsection{Datasets}
\label{sec:datasets}
We evaluate our approach on both conventional CSC datasets and our newly constructed dataset, \textbf{C2EC}.
For conventional CSC, following \citet{li-etal-2024-cllm}, we use two representative datasets: \textbf{CSCD-NS} \cite{hu-etal-2024-cscd} and \textbf{Lemon} \cite{wu-etal-2023-rethinking}.
Both datasets contain text written by native speakers.
CSCD-NS focuses on general domain performance, while Lemon evaluates zero-shot cross-domain capabilities.
The statistics of the datasets are shown in Table~\ref{tab:dataset_statistics}.
As we are unable to access the ECMR-2023 dataset \cite{he-etal-2023-umrspell}, we do not include it in our experiments.

\subsection{Evaluation Metrics}
Following previous works \cite{li-etal-2024-cllm,zhou-etal-2024-simple}, we use character-level correction $F_1$ as our main evaluation metric.
Since sentence-level metrics are widely used in previous works, we also report them in Appendix~\ref{subsec:sentence_level_results}.
Details of the metrics can be found in Appendix~\ref{subsec:evaluation_implementation_and_settings}.

\begin{table*}[tb!]
    \centering
    \renewcommand{\arraystretch}{0.95}
    \setlength{\tabcolsep}{4.0pt}
    \scalebox{0.9}{
        \begin{NiceTabular}{lccccccccc;c;c|>{\columncolor{figure_light_red!6}}c}
            \toprule
            \rowcolor[gray]{.9}
                                                                                            &                               &                                 & \Block[c]{1-9}{\textbf{\textsc{Conventional CSC}}} &                &                &                &                &                &                &                                  &                               & \textbf{\textsc{C2EC}} \\
            \Block[l]{2-1}{\textbf{Model}}                                                  & \Block[c]{2-1}{\textbf{Size}} & \Block[c]{2-1}{\textbf{Method}} & \Block[c]{1-7}{\textbf{Lemon}}                     &                &                &                &                &                &                & \Block[c]{1-1}{\textbf{CSCD-NS}} & \Block[c]{2-1}{\textbf{Avg.}} & \textbf{C2EC}          \\
                                                                                            &                               &                                 & \textit{Car}                                       & \textit{Cot}   & \textit{Enc}   & \textit{Gam}   & \textit{Mec}   & \textit{New}   & \textit{Nov}   & \textit{test}                    &                               & \textit{test}          \\
            \midrule
            \rowcolor[gray]{1.0}
            \Block[c]{1-13}{\textit{Supervised Fine-tuning SoTAs} \cite{li-etal-2024-cllm}} &                               &                                 &                                                                                                                                                                                                                                                      \\
            \midrule
            \texttt{SCOPE}                                                                  & \texttt{0.1B}                 & \texttt{Full}                   & 50.71                                              & 54.89          & 45.23          & 24.74          & 44.44          & 48.72          & 33.17          & 71.70                            & 46.70                         & --                     \\
            \Block[l]{2-1}{\texttt{Qwen1.5}}                                                & \texttt{7B}                   & \texttt{LoRA}                   & 53.87                                              & 58.04          & 54.57          & 37.43          & 61.16          & 60.07          & 41.42          & 71.64                            & 54.77                         & --                     \\
                                                                                            & \texttt{14B}                  & \texttt{LoRA}                   & 57.54                                              & 60.40          & 56.48          & 38.02          & 65.31          & 64.49          & 43.92          & \textbf{73.80}                   & 57.49                         & --                     \\
            \midrule
            \rowcolor[gray]{1.0}
            \Block[c]{1-13}{\textit{Training-free Methods}}                                 &                               &                                                                                                                                                                                                                                                                                        \\
            \midrule
            \texttt{GPT4o-mini}                                                             & \texttt{N/A}                  & \texttt{ICL}                    & 32.13                                              & 29.57          & 41.43          & 12.46          & 34.12          & 33.53          & 26.67          & 40.32                            & 31.28                         & 30.73                  \\
            \texttt{GPT4o}                                                                  & \texttt{N/A}                  & \texttt{ICL}                    & 54.68                                              & 57.25          & 63.02          & 14.60          & 62.92          & 61.37          & 54.10          & 65.43                            & 54.17                         & 45.71                  \\
            \texttt{DeepSeek\,V3}                                                           & \texttt{671B}                 & \texttt{ICL}                    & 60.12                                              & 69.91          & \textbf{68.56} & 38.10          & \textbf{70.34} & 69.67          & \textbf{61.41} & 69.43                            & 63.44                         & 54.62                  \\
            \texttt{DeepSeek\,R1}                                                           & \texttt{671B}                 & \texttt{ICL}                    & 57.80                                              & 63.47          & 62.38          & 45.37          & 69.38          & 68.39          & 58.21          & 62.68                            & 61.61                         & 48.85                  \\
            \hdashedline
            \Block[l]{4-1}{\texttt{Qwen2.5}}                                                & \Block[c]{4-1}{\texttt{7B}}   & \texttt{ICL}                    & 28.98                                              & 39.78          & 36.89          & \wz9.89        & 38.22          & 27.98          & 22.09          & 32.61                            & 29.56                         & 29.03                  \\
                                                                                            &                               & \texttt{ICL-RR}                 & 41.39                                              & 55.61          & 49.23          & 24.13          & 51.19          & 44.48          & 34.14          & 54.20                            & 44.30                         & 41.01                  \\
                                                                                            &                               & \texttt{TfPf}                   & 55.25                                              & 64.31          & 53.89          & 41.08          & 57.47          & 63.39          & 45.53          & 62.45                            & 55.42                         & 41.50                  \\
                                                                                            &                               & \texttt{OUR}                    & 61.80                                              & \textbf{71.05} & 65.86          & 51.67          & 68.65          & 69.34          & 51.89          & 71.03                            & 63.91                         & 56.02                  \\
            \hdashedline
            \Block[l]{4-1}{\texttt{Qwen2.5}}                                                & \Block[c]{4-1}{\texttt{14B}}  & \texttt{ICL}                    & 35.93                                              & 49.65          & 42.04          & 26.45          & 45.37          & 38.02          & 31.45          & 38.39                            & 38.41                         & 32.52                  \\
                                                                                            &                               & \texttt{ICL-RR}                 & 48.80                                              & 55.39          & 52.63          & 40.02          & 53.67          & 55.46          & 44.43          & 55.68                            & 50.76                         & 42.86                  \\
                                                                                            &                               & \texttt{TfPf}                   & 55.51                                              & 62.50          & 54.43          & 37.90          & 56.58          & 64.25          & 46.74          & 62.53                            & 55.06                         & 40.96                  \\
                                                                                            &                               & \texttt{OUR}                    & \textbf{64.62}                                     & 70.81          & 68.50          & \textbf{51.92} & 68.24          & \textbf{71.85} & 53.68          & 72.75                            & \textbf{65.30}                & \textbf{57.41}         \\
            \bottomrule
        \end{NiceTabular}
    }
    \caption{
        Comparison between our method and the baseline methods on conventional CSC datasets.
    }
    \label{tab:main_results}
\end{table*}

\subsection{Baselines}
We compare our approach against three training-free baselines:
\begin{inparaenum}[\itshape a)]
    \item \textbf{In-context Learning} (\texttt{ICL}): This method prompts LLMs with 10 exemplars (5 erroneous, 5 correct sentences) randomly selected and shuffled for each input. During inference, we use beam search with the same beam size as our approach;\footnote{For conventional CSC datasets, exemplars are randomly sampled from the CSCD-NS training set, while for C2EC, they come from its development set. Due to the random nature of \texttt{ICL}, we report the results averaged across 3 runs.}
    \item \textbf{Training-free Prompt-free Method} (\texttt{TfPf}, \citet{zhou-etal-2024-simple}); and
    \item \textbf{ICL with Reranking} (\texttt{ICL-RR}): This hybrid method first generates $K$ candidates using \texttt{ICL}, then reranks them using an extended \texttt{TfPf} model that supports insertion and deletion operations.
\end{inparaenum}

For reference, we also include \texttt{ICL} results from leading LLMs accessed via API:\footnote{Due to API constraints and cost, instead of beam search, greedy decoding (\texttt{temperature=0.0}) was used, and single-run results are reported.} \texttt{GPT4o-mini}\rlap{,}\footnote{Version: \texttt{gpt-4o-mini-2024-07-18}} \texttt{GPT4o} \cite{hurst-etal-2024-gpt4o}\rlap{,}\footnote{Version: \texttt{gpt-4o-2024-11-20}} and the 671B-parameter models \texttt{DeepSeek\,V3} and \texttt{R1}\footnote{Versions: \texttt{DeepSeek\,V3} (released on 2024-12-26) and \texttt{R1} (released on 2025-01-20).}\cite{deepseekai-2024-deepseek-v3,deepseek-r1-2025}.

Additionally, the supervised fine-tuning (\texttt{SFT}) results on conventional CSC datasets from \citet{li-etal-2024-cllm} are also included for better understanding.

\paragraph{Training Details of SFT Baselines}
The \texttt{SFT} models were trained on a combined dataset consisting of 271k pseudo sentence pairs \cite{wang-etal-2018-hybrid} and the \textbf{CSCD-NS} training data.
This means that while CSCD-NS is an in-domain evaluation, the Lemon dataset serves as a cross-domain dataset for evaluating the generalization capabilities.

\subsection{Model Selection}
We use the \texttt{Qwen2.5} model series \cite{yang-etal-2024-qwen25} as the main model for our experiments, as it is one of the most recent open-source LLMs with strong Chinese language capabilities.

For \texttt{ICL} experiments, we use the ``\texttt{Instruct}'' version, which is optimized for instruction following.
As for our approach, we use the ``\texttt{Base}'' version.
To reduce the computational cost, we use \texttt{7B} variant for detailed analysis.

To further validate the effectiveness of our approach, three other LLM families, \texttt{Qwen1.5}, \texttt{Baichuan2} and \texttt{InternLM2}, are also discussed in the discussion section.

\subsection{Hyperparameters}
We largely follow the hyperparameter settings in \citet{zhou-etal-2024-simple}, using a beam size of 8 and $\alpha$ for length reward of 2.5.
Through development set tuning, we set the weights for insertion and deletion as 8.5 and 9.0, respectively, and the temperature of the prompt-based scoring as 1.5.
During the ablation study, we tune the temperature for each setting to maximize its respective $F_1$ score.

    \begin{table*}[tb!]
    \centering
    \scalebox{0.9}{
        \begin{NiceTabular}{ccccc;ccc;ccc}
            \toprule
            \rowcolor[gray]{1.0}
                       & \Block[c]{2-1}{\texttt{Prompt}                                                                                                                                                                                                                                                         \\[-2pt]\texttt{Template}}         & \Block[c]{2-1}{\texttt{Prompt}\\[-2pt]\texttt{LLM}}          & \Block[c]{2-1}{\texttt{Pure}\\[-2pt]\texttt{LLM}}         & \Block[c]{2-1}{\texttt{Distance}\\[-2pt]\texttt{Metric}}      & \Block[c]{1-3}{\textbf{CSCD-NS} \textit{dev}} &                &                & \Block[c]{1-3}{\textbf{C2EC} \textit{dev}} &                &                \\
                       &                                            &                                            &                                            &                                           & \textbf{P}     & \textbf{R}     & \textbf{F$_1$} & \textbf{P}     & \textbf{R}     & \textbf{F$_1$} \\
            \midrule
            \rowcolor[gray]{0.98}
            \texttt{1} & \texttt{Minimal}                           & \texttt{Base}                              & \texttt{Base}                              & \texttt{Levenshtein}                      & 67.70          & 73.69          & 70.57          & \textbf{65.14} & 49.18          & 56.05          \\
            \midrule
            \texttt{2} & \texttt{Minimal}                           & \texttt{Base}                              & \texttt{Base}                              & \textcolor{figure_blue}{\texttt{Hamming}} & \textbf{72.49} & \textbf{74.90} & \textbf{73.68} & 65.06          & 41.71          & 50.83          \\
            \texttt{3} & \textcolor{figure_blue}{--}                & \textcolor{figure_blue}{--}                & \texttt{Base}                              & \texttt{Levenshtein}                      & 60.04          & 65.23          & 62.53          & 54.80          & 37.43          & 44.48          \\
            \texttt{4} & \texttt{Minimal}                           & \texttt{Base}                              & \textcolor{figure_blue}{--}                & \texttt{Levenshtein}                      & 61.12          & 57.91          & 59.47          & 55.96          & 46.18          & 50.60          \\
            \hdashedline
            \texttt{5} & \textcolor{figure_blue}{\texttt{Detailed}} & \texttt{Base}                              & \texttt{Base}                              & \texttt{Levenshtein}                      & 67.60          & 71.65          & 69.57          & 65.12          & 45.90          & 53.85          \\
            \texttt{6} & \texttt{Minimal}                           & \textcolor{figure_blue}{\texttt{Instruct}} & \textcolor{figure_blue}{\texttt{Instruct}} & \texttt{Levenshtein}                      & 58.35          & 69.50          & 63.44          & 55.04          & 50.27          & 52.55          \\
            \texttt{7} & \textcolor{figure_blue}{\texttt{Detailed}} & \textcolor{figure_blue}{\texttt{Instruct}} & \textcolor{figure_blue}{\texttt{Instruct}} & \texttt{Levenshtein}                      & 67.33          & 68.44          & 67.88          & 62.21          & \textbf{51.73} & 56.49          \\
            \bottomrule
        \end{NiceTabular}
    }
    \caption{
        Ablation study on different parts of our approach.
    }
    \label{tab:ablation:parts}
\end{table*}

\section{Main Results}
As shown in Table \ref{tab:main_results}, our method performs better than \texttt{ICL}, \texttt{ICL-RR}, and \texttt{TfPf} on both conventional CSC datasets and the C2EC dataset.
Compared to \texttt{TfPf}, from which our method is extended, we achieve improvements of 8.49 and 10.24 on average with 7B and 14B models, respectively.
Compared to \texttt{ICL-RR}, our method is shown to be a better way to combine the advantages of prompt-based LLMs and \texttt{TfPf}.
This is because \texttt{ICL-RR} can only choose from the top $K$ candidates from \texttt{ICL}.
If none of these candidates are good, reranking cannot improve the final result.

Compared to \texttt{SFT} models from \citet{li-etal-2024-cllm}, which are trained on the training set of the CSCD-NS dataset, our method shows better performance on the \textbf{out-of-domain} Lemon dataset.
This indicates that \texttt{SFT} methods may not generalize well to new data they have not seen during training.
It is also worth noting that, since the \texttt{SFT} models were trained with the \texttt{Qwen1.5} series, which could make a direct comparison with our method unfair, we also provide our results using the \texttt{Qwen1.5} series in Appendix~\ref{app:supervised_fine_tuning}.

Compared to recent leading LLMs (e.g., the 671B parameter \texttt{DeepSeek\,V3}), our method enables much smaller 7B and 14B models to be on par with them without any training.

For a better understanding of the performance of our method, we also provide several qualitative results in Appendix~\ref{subsec:qualitative_analysis}.

An interesting phenomenon worth noting is that the reasoning model \texttt{DeepSeek\,R1} shows lower performance than \texttt{DeepSeek\,V3}, its non-reasoning variant, on both CSC and C2EC tasks.
We find this may be caused by incorrect reasoning.
More discussion of this is provided in Appendix~\ref{subsec:incorrect_thinking_may_lead_to_wrong_corrections}.

    \section{Discussion}
\label{sec:analysis:finetuning}

\subsection{Hamming Distance vs. Levenshtein Distance}
Recall that in this work, we extend \texttt{TfPf} in two ways.
The first is replacing the original Hamming distance with a Levenshtein distance to allow for insertion and deletion operations.

The second row of Table~\ref{tab:ablation:parts} shows the results when we revert to the original Hamming distance.
While the Levenshtein distance-based model performs better on the C2EC dataset, it degrades the performance on the CSCD-NS dataset, the conventional CSC dataset.

This can be attributed to two factors.
First, the conventional CSC dataset only contains substitution errors; any insertion or deletion will lead to an over-correction.
Second, insertion and deletion operations may compete with replacement operations, leading to mis-corrections.
For example, when deletion is allowed, the model may incorrectly delete a misspelled character instead of replacing it with the correct one.

\begin{figure}[tb!]
    \centering
    \captionsetup[subfigure]{skip=-1pt, margin=5pt}
    \begin{tikzpicture}[
            legend/.style={
                    fill=white,
                    font=\footnotesize,
                    inner sep=2pt,
                    minimum width=1.0cm,
                    text opacity=1.0,
                    fill opacity=1.0,
                },
            diff label/.style={
                    font=\scriptsize,
                    inner sep=0.5pt,
                    outer sep=1.5pt,
                    fill=white,
                    fill opacity=0.9,
                    text opacity=1.0,
                    rounded corners=1pt,
                },
            trim left
        ]
        \centering
        \begin{groupplot}[
                group style={
                        group size=1 by 2,
                        x descriptions at=edge bottom,
                        horizontal sep=0.7cm,
                        vertical sep=0.1cm,
                    },
                width=1.05\linewidth,
                height=0.65\linewidth,
                xlabel={Model Size (B)},
                xmin=0,
                xmax=15,
                xmajorgrids=true,
                xtick={0.5, 1.5, 3, 7, 14},
                xticklabels={0.5, 1.5, 3, 7, 14},
                /tikz/font=\footnotesize,
                ylabel shift=-4pt,
                xlabel shift=-4pt,
                yticklabel shift=-2pt,
                xticklabel shift=-1pt,
                legend style={
                        at={(0.985,0.03)},
                        anchor=south east,
                        font=\footnotesize,
                    },
                legend columns=2,
            ]
            \nextgroupplot[ymin=0,ymax=85]
            \addplot+ [mark=triangle*, draw=black, thick,
                densely dashed,
                mark size=2.5pt,
                mark options={fill=white, fill opacity=1.0, solid},
                opacity=1.0,
            ] table [row sep=\\] {
                    x	y\\
                    0.5	1.39\\
                    1.5	18.56\\
                    3	26.75\\
                    7	31.65\\
                    14	39.68\\
                };
            \addplot+ [mark=*, draw=black, thick,
                mark size=1.25pt,
                mark options={fill=white, fill opacity=1.0, solid},
                opacity=1.0,
            ] table [row sep=\\] {
                    x	y\\
                    0.5	8.61\\
                    1.5	47.81\\
                    3	49.75\\
                    7	52.59\\
                    14	55.68\\
                };
            \addplot+ [mark=square*, draw=figure_red, thick,
                dashed,
                mark size=1.25pt,
                mark options={fill=white, fill opacity=1.0, solid},
                opacity=1.0,
            ] table [row sep=\\] {
                    x	y\\
                    0.5	56.04\\
                    1.5	58.82\\
                    3	59.11\\
                    7	61.11\\
                    14	59.93\\
                };
            \addplot+ [mark=*, draw=figure_blue, thick,
                mark size=1.25pt,
                mark options={fill=figure_blue, fill opacity=1.0, solid},
                opacity=1.0,
            ] table [row sep=\\] {
                    x	y\\
                    0.5	60.46\\
                    1.5	65.78\\
                    3	68.39\\
                    7	70.57\\
                    14	72.77\\
                };
            \nextgroupplot[ymin=0,ymax=65]
            \addplot+ [mark=*, draw=figure_blue, thick,
                mark size=1.25pt,
                mark options={fill=figure_blue, fill opacity=1.0, solid},
                opacity=1.0,
            ] table [row sep=\\] {
                    x	y\\
                    0.5	41.55\\
                    1.5	50.23\\
                    3	52.42\\
                    7	56.05\\
                    14	57.64\\
                };
            \addplot+ [mark=square*, draw=figure_red, thick,
                dashed,
                mark size=1.25pt,
                mark options={fill=white, fill opacity=1.0, solid},
                opacity=1.0,
            ] table [row sep=\\] {
                    x	y\\
                    0.5	42.11\\
                    1.5	40.36\\
                    3	38.99\\
                    7	40.16\\
                    14	39.80\\
                };

            \addplot+ [mark=*, draw=black, thick,
                mark size=1.25pt,
                mark options={fill=white, fill opacity=1.0, solid},
                opacity=1.0,
            ] table [row sep=\\] {
                    x	y\\
                    0.5	4.94\\
                    1.5	32.49\\
                    3	28.95\\
                    7	38.39\\
                    14	40.00\\
                };
            \addplot+ [mark=triangle*, draw=black, thick,
                densely dashed,
                mark size=2.5pt,
                mark options={fill=white, fill opacity=1.0, solid},
                opacity=1.0,
            ] table [row sep=\\] {
                    x	y\\
                    0.5	2.38\\
                    1.5	13.02\\
                    3	17.08\\
                    7	27.54\\
                    14	28.59\\
                };
            \legend{\texttt{OUR}, \texttt{TfPf}, \texttt{ICL-RR}, \texttt{ICL}}
        \end{groupplot}
        \node[anchor=north, legend] at (group c1r1.north) {\textbf{CSCD-NS} \textit{dev}};
        \node[anchor=north, legend] at (group c1r2.north) {\textbf{C2EC} \textit{dev}};
    \end{tikzpicture}
    \caption{
        Results of different model sizes.
    }
    \label{fig:model_size}
\end{figure}

\subsection{Effectiveness of Dual-LLM}
\label{sec:analysis:prompt}
The second extension is incorporating a prompt-based LLM probability into the \texttt{TfPf} framework.
The third row of Table~\ref{tab:ablation:parts} shows the results when we remove the prompt-based LLM probability.
We observe large performance drops in both precision and recall on both datasets, indicating that the prompt-based LLM probability is a crucial component of our method.
An interesting question is \textit{why we still need the pure LLM probability when we already have the prompt-based LLM probability}.
The fourth row of Table~\ref{tab:ablation:parts} shows the results when we remove the pure LLM probability.
Removing it leads to performance drops of 11.10 and 5.45 points on the CSCD-NS and C2EC datasets, respectively.
As pointed out in the reinforcement learning perspective in Appendix~\ref{sec:reinforcement_learning_perspective}, the pure LLM probability evaluates fluency and correctness from a pure language model perspective, encouraging the model to generate more fluent and correct sentences.

\begin{table*}[tb!]
    \centering
    \scalebox{0.9}{
        \begin{NiceTabular}{lcccc;ccc}
            \toprule
            \rowcolor[gray]{1.0}
            \Block[l]{2-1}{\textbf{Model}} & \Block[c]{2-1}{\phantom{aa}\textbf{Method}\phantom{aa}} & \Block[c]{1-3}{\textbf{CSCD-NS} \textit{dev}} &                &                & \Block[c]{1-3}{\textbf{C2EC} \textit{dev}} &                &                \\
                                           &                                                         & \textbf{P}                                    & \textbf{R}     & \textbf{F$_1$} & \textbf{P}                                 & \textbf{R}     & \textbf{F$_1$} \\
            \midrule
            \Block[l]{4-1}{\texttt{Qwen2.5 7B}}
                                           & \texttt{ICL}                                            & 24.11                                         & 46.07          & 31.65          & 20.55                                      & 41.74          & 27.54          \\
                                           & \texttt{ICL-RR}                                         & 47.83                                         & 58.39          & 52.59          & 33.24                                      & 45.45          & 38.39          \\
                                           & \texttt{TfPf}                                           & 57.98                                         & 64.61          & 61.11          & 52.03                                      & 32.70          & 40.16          \\
                                           & \texttt{OUR}                                            & \textbf{67.70}                                & \textbf{73.69} & \textbf{70.57} & \textbf{65.14}                             & \textbf{49.18} & \textbf{56.05} \\
            \midrule
            \Block[l]{4-1}{\texttt{Qwen1.5 7B}}
                                           & \texttt{ICL}                                            & \wz9.32                                       & 43.12          & 15.33          & \wz7.40                                    & 40.56          & 12.51          \\
                                           & \texttt{ICL-RR}                                         & 18.50                                         & 53.68          & 27.51          & 10.59                                      & 44.87          & 17.13          \\
                                           & \texttt{TfPf}                                           & 52.42                                         & 60.69          & 56.25          & 51.05                                      & 31.06          & 38.62          \\
                                           & \texttt{OUR}                                            & \textbf{62.77}                                & \textbf{71.03} & \textbf{66.64} & \textbf{62.65}                             & \textbf{43.08} & \textbf{51.05} \\
            \hdashedline
            \Block[l]{4-1}{\texttt{Baichuan2 7B}}
                                           & \texttt{ICL}                                            & 12.41                                         & 36.36          & 18.49          & \wz7.79                                    & 26.02          & 11.99          \\
                                           & \texttt{ICL-RR}                                         & 27.65                                         & 48.42          & 35.19          & 15.05                                      & 29.90          & 20.01          \\
                                           & \texttt{TfPf}                                           & 67.96                                         & \textbf{62.96} & 65.37          & 58.21                                      & 29.69          & 39.32          \\
                                           & \texttt{OUR}                                            & \textbf{70.87}                                & 61.16          & \textbf{65.66} & \textbf{62.23}                             & \textbf{34.06} & \textbf{44.03} \\
            \hdashedline
            \Block[l]{4-1}{\texttt{InternLM2 7B}}
                                           & \texttt{ICL}                                            & 22.16                                         & 39.47          & 28.38          & 15.56                                      & 30.90          & 20.43          \\
                                           & \texttt{ICL-RR}                                         & 24.97                                         & 39.83          & 30.70          & 23.74                                      & 31.75          & 27.17          \\
                                           & \texttt{TfPf}                                           & 61.12                                         & 63.08          & 62.08          & 49.92                                      & 26.78          & 34.86          \\
                                           & \texttt{OUR}                                            & \textbf{65.71}                                & \textbf{68.29} & \textbf{66.97} & \textbf{60.59}                             & \textbf{39.62} & \textbf{47.91} \\
            \bottomrule
        \end{NiceTabular}
    }
    \caption{
        Results of applying our method to different LLM families.
        All models are with 7B parameters.
    }
    \label{tab:analysis:other_lm}
\end{table*}

\subsection{Impact of Different Prompt Templates}
\label{sec:analysis:prompt_template}
In Section~\ref{sec:prompt_template}, we introduce two prompt templates:
a minimal prompt and a detailed prompt with more instructions.
This section investigates the impact of different prompt templates.
Rows 5-7 of Table~\ref{tab:ablation:parts} show the results of using the minimal prompt and the detailed prompt with the base LLM and instruction-tuned LLM.
The results show that base LLM favors the minimal prompt, while instruction-tuned versions prefer the longer and more complex prompt.
Compared to instruction-tuned versions, base LLM is more robust to prompt variations.

\subsection{Impact of the LLM Size}
\label{sec:ablation:size}
To investigate the impact of model size, we conduct experiments across different sizes of the Qwen2.5 series.
The results are shown in Figure~\ref{fig:model_size}.
We observe that our method's performance generally improves with increasing model size.
In contrast, \texttt{TfPf}'s performance does not consistently improve,
indicating that our method better leverages the scale of larger LLMs.
However, on the 0.5B model, which has very limited prompt understanding, \texttt{TfPf} may outperform our method.

\subsection{Impact of Different LLM Families}
\label{app:other_llm}
To investigate if our method can improve the performance of other LLMs, we conducted experiments on three additional LLMs: \texttt{Qwen1.5} \cite{bai-etal-2023-qwen}, \texttt{Baichuan2} \cite{yang-etal-2023-baichuan}, and \texttt{InternLM2} \cite{cai-etal-2024-internlm2}.
For these LLMs, we used the 7B variant.

The results are shown in Table~\ref{tab:analysis:other_lm}.
Our method consistently improves the performance of these LLMs.
However, the performance gain with \texttt{TfPf} varies across different LLMs.
We observed that the performance improvement might be related to the recall value of \texttt{ICL} for different LLMs.
A higher recall value of \texttt{ICL} corresponds to a larger performance improvement with our method.
We plan to explore the underlying mechanism of this phenomenon in the future.

\subsection{Performance on Different Error Types}
\label{sec:analysis:error_type}
Figure~\ref{fig:error_edit_types} shows the performance across different error types.
Our method outperforms baselines on all error types.
The difficulty levels of different error types vary.
Substitution (misspelling) errors are the easiest to correct, with all methods achieving relatively high performance.
Redundant errors are the most challenging.
This is mainly because models tend to add optional characters to the error-free sentence, leading to over-correction.
An example of this is given in Appendix~\ref{subsec:qualitative_analysis}.

\begin{figure}[tb!]
    \centering
    \begin{tikzpicture}[trim axis left]
        \begin{axis}[
                ybar,
                width=1.05\linewidth,
                height=0.75\linewidth,
                bar width=10pt,
                enlarge x limits=0.25,
                symbolic x coords={
                        MIS, RED, SUB
                    },
                xtick=data,
                /tikz/font=\footnotesize,
                ymin=0,
                ymax=65,
                ylabel shift=-4pt,
                xlabel shift=-4pt,
                yticklabel shift=-2pt,
                xticklabel shift=-1pt,
                legend style={
                        at={(0.5,0.97)},
                        anchor=north,
                        font=\footnotesize,
                    },
                legend columns=4,
            ]

            \addplot[
                draw=black,
                thick,
            ] coordinates {
                    (MIS, 18.30) (RED, 20.69) (SUB, 31.20)
                };

            \addplot[
                pattern=north west lines,
                thick,
            ] coordinates {
                    (MIS, 28.25) (RED, 24.70) (SUB, 46.42)
                };

            \addplot[
                pattern=crosshatch dots,
                thick,
                pattern color=figure_red,
                draw=black
            ] coordinates {
                    (MIS, 0) (RED, 0) (SUB, 47.931)
                };

            \addplot[
                fill=figure_blue!80,
                thick,
                draw=black
            ] coordinates {
                    (MIS, 47.663) (RED, 31.428) (SUB, 60.679)
                };

            \draw [
                densely dashed,
                gray,
                thin
            ] (rel axis cs:0.333,0) -- (rel axis cs:0.333,1);

            \draw [
                densely dashed,
                gray,
                thin
            ] (rel axis cs:0.666,0) -- (rel axis cs:0.666,1);

            \legend{\texttt{ICL}, \texttt{ICL-RR}, \texttt{TfPf},\texttt{OUR}}
        \end{axis}
    \end{tikzpicture}
    \caption{%
        The results of different types of error on the C2EC dev set.
        MIS: Missing, RED: Redundant, SUB: Substitution (Misspelling)
    }%
    \label{fig:error_edit_types}
\end{figure}

\subsection{More Discussion}
Due to space limitations, some interesting discussions are included in Appendix~\ref{app:more_discussion}.
These include:
\begin{inparaenum}[\itshape a)]
    \item Investigation of beam size impact (\ref{app:beam_size}),
    \item Ablation study on length reward and faithfulness reward adopted from \texttt{TfPf} (\ref{app:ablation:two_reward}),
    \item A fair comparison with supervised fine-tuning on \texttt{Qwen1.5} series (\ref{app:supervised_fine_tuning})
    \item How to adjust the precision-recall trade-off (\ref{app:precision_recall_trade_off}), and
    \item Runtime analysis (\ref{app:runtime}).
\end{inparaenum}

    \section{Related Works}
\label{sec:related_work}

\subsection{Datasets of Chinese Spelling Correction}
Chinese Spelling Correction (CSC) has long been an important research area in NLP, with new datasets continuously being developed to address various needs.
The Sighan series \cite{wu-etal-2013-chinese,yu-etal-2014-overview,tseng-etal-2015-introduction} is one of the earliest and most influential collections of CSC datasets.
While widely used, they are criticized for their poor annotation quality, limited domain, and unrealistic error patterns \cite{yang-etal-2023-chinese,wu-etal-2023-rethinking,sun-etal-2024-two}.

While researchers like \cite{yang-etal-2023-chinese,sun-etal-2024-two} have tried to improve Sighan datasets through re-annotation, the inherent issues of unrealistic domain and limited error patterns remain.
To address these issues, researchers manually created errors across financial, official documents and medical domains \cite{jiang-etal-2022-mcscset,lv-etal-2023-ecspell}.
Instead of creating errors from correct sentences, \citet{wu-etal-2023-rethinking} collected and annotated real errors from seven different domains, building a large dataset that challenges models by not providing training data. Similarly, \citet{hu-etal-2024-cscd} created a new dataset with real errors found on social media.

Some works have tried to broaden the scope of conventional CSC datasets.
Similar to our work, \citet{he-etal-2023-umrspell} also pointed out that existing datasets mainly focus on substitution errors, overlooking two other common types: insertion and deletion errors, limiting the practicality of the task.
To address this, they created the ECMR-2023 dataset by randomly adding, removing, or replacing characters in correct sentences.
However, these artificially generated errors may not reflect real-world mistakes well.
In contrast, we build our C2EC dataset using existing Chinese text correction data, providing a more realistic benchmark for general Chinese character error correction.

\subsection{Approaches of Chinese Spelling Correction}
For many years, BERT-based models have dominated CSC research \cite{zhang-etal-2020-spelling,xu-etal-2021-read,zhu-etal-2022-mdcspell,liang-etal-2023-disentangled}.
While these models show strong performance on in-domain datasets, recent studies reveal their limitations when applied to different domains \cite{wu-etal-2023-rethinking,liu-etal-2024-rephrasing}.

With the advent of LLMs, researchers have begun exploring their potential for CSC.

Early attempts focused on prompt-based methods \cite{li-etal-2023-ineffectiveness}.
For instance, \citet{dong-etal-2024-rich} enhanced prompts with pronunciation and glyph information.
However, lightweight LLMs, such as those with 7B or 14B parameters, still struggle to achieve satisfactory performance with prompt-based methods.
Compared to prompt-based methods, supervised fine-tuning methods have shown more effectiveness.
A representative work is \citet{li-etal-2024-cllm}, which introduced a novel training paradigm that retrains LLMs at the character level to better align with CSC requirements.

Recently, \citet{zhou-etal-2024-simple} proposed \texttt{TfPf}, a training-free and prompt-free framework that achieves comparable cross-domain performance to \texttt{SFT} methods.
As detailed in \S\ref{sec:baseline_tfpf_approach}, \texttt{TfPf} combines a language model with a distance metric to balance fluency and minimal edits.
Our work builds upon this elegant framework by improving its performance and extending it to handle the C2EC task.

    \section{Conclusion}
In this work, we propose the task of C2EC, which is an extended CSC task that handles misspellings and two previously ignored errors: redundant and missing characters.
To support this task, we construct a focused C2EC dataset containing real-world errors by combining and manually verifying two existing datasets.
We propose a training-free method for the C2EC task by extending the \texttt{TfPf} method.
Experiments show that our method achieves large improvements over both 10-shot in-context learning and the \texttt{TfPf} baseline on conventional CSC and C2EC datasets, achieving competitive performance with models nearly 50 times larger.

    \section*{Limitations}
There are several limitations in this work that we plan to investigate and address in future work.

\paragraph{Research Scope}
The scope of this study is limited to Chinese character error correction. However, we believe that with some modifications, our approach can be applied to other languages and more complex tasks like grammatical error correction and sentence simplification.
We plan to extend our approach to these areas in future work.

In this paper, we focus on evaluating LLM performance during inference under zero-shot or few-shot scenarios.
When sufficient annotated data is available, how to effectively utilize it to improve performance of our approach is an interesting question we plan to explore in future work.

Moreover, limited by computational resources, we have not tested our approach on larger models.
We believe that models with more parameters would also benefit from our approach.

\paragraph{Speed and Resources}
Our approach relies on large language models, which inherently demand significant computational resources.
Additionally, the inference speed is constrained by the requirement of two forward passes of the LLM at each generation step.
Optimizing the implementation, for instance, by ensuring compatibility with modern LLM inference frameworks, might leverage the latest advancements in LLM inference to improve the speed of our approach.

\paragraph{Flexibility}
Our approach requires access to the model's probability distribution, which makes it unable to be directly applied to API-accessed models that are often more powerful.
However, given the flexibility of prompt-based LLMs, we believe there is potential to leverage results from API-accessed models to further improve performance.

\section*{Ethics Statement}
Our proposed dataset is built upon existing publicly available datasets. We have properly cited the original datasets in our paper and ensured that our use is consistent with their original intent.
For works that use our dataset, we require them to appropriately cite the original datasets.

For this work, we manually verified the dataset to ensure quality.
We recruited four graduate students who are native Chinese speakers with high Chinese proficiency.
The verification process took about 16 working hours per annotator.
Each annotator was compensated at a rate of ¥25 per hour.

    \ifarxiv%
        \section*{Acknowledgments}
First of all, we would like to express our sincere gratitude to the anonymous reviewers for their valuable suggestions and comments.

We sincerely thank Haozhe Zhou, Ziheng Qiao, Haochen Jiang, and Yumeng Liu for their manual verification of the dataset. We are also deeply grateful to Chen Gong for her continuous support throughout this research.

This work was supported by National Natural Science Foundation of China (Grant No. 62261160648 and 62176173) and a project funded by the Priority Academic Program Development (PAPD) of Jiangsu Higher Education Institutions.

    \fi%

    \bibliography{custom}

    \clearpage
    \appendix

\section{Details of C2EC Dataset}
\subsection{Annotation UI for Data Verification}
\label{sec:anno_ui}
\begin{figure}[tb!]
    \centering
    \includegraphics[width=1.0\linewidth]{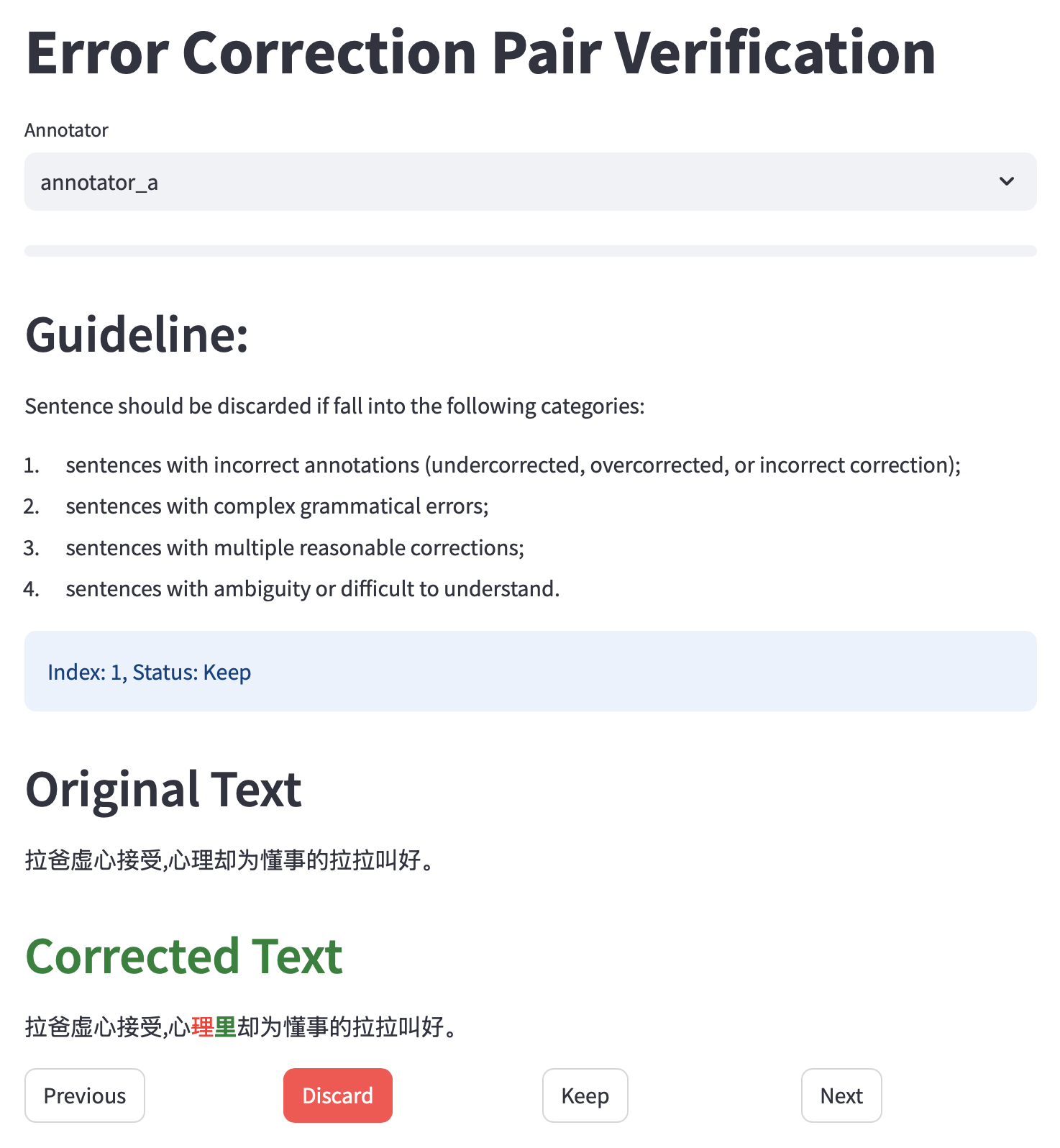}
    \caption{
        The annotation UI for data verification.
    }
    \label{fig:anno_ui}
\end{figure}

Figure~\ref{fig:anno_ui} shows our verification user interface.
To maintain verification quality, we always present the guidelines to the annotators at the top of the UI.
Additionally, we highlight the differences between the original and corrected sentences to help the annotators easily identify the changes.
We also require the annotators to carefully remove any sentences containing sensitive or offensive content.
However, we did not find any such sentences during verification.

\subsection{Discarded Cases}
\label{sec:discard_cases}
\begin{table}[t!]
    \renewcommand{\arraystretch}{1.1}
    \centering
    {
        \scalebox{0.9}{
            \begin{NiceTabular}{p{1.5cm}p{6.0cm}}
                \toprule
                \Block{1-2}{\textit{Incorrect annotations}}                          &                                                  \\
                \midrule
                \rowcolor[gray]{.95}
                Input                                                                & 特种兵那么厉害，一旦犯罪了怎么{}\wrong{组织}？其实国家早有防备             \\
                Annotation                                                           & 特种兵那么厉害，一旦犯罪了怎么{}\wrong{组}{}\correct{止}？其实国家早有防备 \\
                \midrule
                \midrule
                \Block{1-2}{\textit{Having complex grammatical errors}}              &                                                  \\
                \midrule
                \rowcolor[gray]{.95}
                Input                                                                & 千佛山起源于济南南部山区，山势宛如\wrong{巨大的一轮}秋月，弯伏恢宏。           \\
                Annotation                                                           & 千佛山起源于济南南部山区，山势宛如\correct{一轮巨大的}秋月，弯伏恢宏。         \\
                \midrule
                \midrule
                \Block{1-2}{\textit{Having multiple reasonable corrections}}         &                                                  \\
                \midrule
                \rowcolor[gray]{.95}
                Input                                                                & 二人此时此刻，站在一起，令\wrong{人}四周不少人，心中羡慕。                \\
                Annotation                                                           & 二人此时此刻，站在一起，令\correct{得}四周不少人，心中羡慕。              \\
                \midrule
                \midrule
                \Block{1-2}{\textit{Ambiguous or difficult to understand sentences}} &                                                  \\
                \midrule
                \rowcolor[gray]{.95}
                Input                                                                & 薛静妃点头，“我想要带着整个阴阳宗去混沌\wrong{妖族}！”                 \\
                Annotation                                                           & 薛静妃点头，“我想要带着整个阴阳宗去混沌\correct{宇宙}！”               \\
                \bottomrule
            \end{NiceTabular}
        }
    }
    \caption{
        Some examples of the sentences that are discarded during the manual verification.
    }
    \label{tab:discard_cases}
\end{table}

We present examples of sentences discarded during manual verification in Table~\ref{tab:discard_cases}.
The first example involves a sentence with incorrect annotation.
The input sentence incorrectly spells the word ``阻止'' (prevent, \textit{zǔ zhǐ}) as ``组织'' (organize, \textit{zǔ zhǐ}), but the annotator from the original dataset only corrected the character to ``止'', leaving ``组'' unchanged.
The second example is a sentence with complex grammatical errors.
In Chinese textbooks, there is a recommended rule for arranging multiple adjectives in a sentence for a formal style.
For instance, adjectives indicating time or location should precede those of quantity, and quantity adjectives should precede attributive adjectives.
Thus, a grammatically correct phrase in this context is ``一轮巨大的秋月''.
However, correcting such sentences is beyond the scope of this work.
The third example is a sentence with multiple reasonable corrections.
The incorrect character ``人'' (person, \textit{rén}) can be corrected by either removing it or changing it to ``得'' (an auxiliary verb, \textit{dé}).
The fourth example is an ambiguous sentence that requires additional context for accurate correction.
Given the context, we cannot determine whether the term ``混沌妖族'' (\textit{Chaos Demon Clan}, \textit{hùn dùn yāo zú}) is valid within the novel, or if it should be corrected to another term, leading us to discard this sentence.

We plan to re-annotate discarded sentences in the future and share them with the original authors to improve the dataset.

\section{Distance Metrics}
\label{sec:distance_metric}

\subsection{Hamming Distance}
\label{subsec:hamming_distance}
Hamming distance measures the position-wise differences between two strings of equal length:
\begin{equation}
    \mathtt{Dist}(\substring{x}, \substring{y}) = \sum_{i=1}^{n} w(x_i, y_i)
\end{equation}
where $\substring{x}$ and $\substring{y}$ are strings of the same length $n$, and $w(x_i, y_i)$ is the weight assigned to the characters $x_i$ and $y_i$.
In the standard scenario, $w(x_i, y_i) = 1$ if $x_i \neq y_i$ and $0$ otherwise.
For instance, the Hamming distance between \texttt{a\correct{b}c\correct{de}f} and \texttt{a\correct{1}c\correct{23}f} is~3.

\paragraph{Distortion Model of \texttt{TfPf} as a Weighted Hamming Distance}
\label{subsec:distortion_model_of_tfpf_as_a_weighted_hamming_distance}
The distortion model $\log p_{\mathtt{DM}}(\boldsymbol{x}\mid \boldsymbol{y})$ in \texttt{TfPf} is character-level factorized:
\begin{equation}
    \log p_{\mathtt{DM}}(\boldsymbol{x}\mid \boldsymbol{y}) = \sum_{i=1}^{n} \log p_{\mathtt{DM}}(x_i\mid y_i)
\end{equation}
By interpreting $- \log p_{\mathtt{DM}}(x_i\mid y_i)$ as the weight for characters $x_i$ and $y_i$, the distortion model of \texttt{TfPf} can be seen as a weighted Hamming distance.

\subsection{Levenshtein Distance}
Levenshtein distance measures the difference between two strings, which may have different lengths, by calculating the minimum number of edit operations (\texttt{substitution}, \texttt{insertion}, and \texttt{deletion}) required.
For example, the Levenshtein distance between \texttt{abc\correct{d}e} and \texttt{a\correct{1}bc\correct{2}e} is 2, including an insertion of `1' between `a' and `b', and a substitution of `d' with `2'.

The Levenshtein distance can be computed using a dynamic programming algorithm with the following recurrence relation:

\begin{equation*}
    \begin{aligned}
         & \mathtt{Dist}(\substring{x}_{\le m}, \substring{y}_{\le n})                                                \\
         & \ = \min \begin{cases}
                        \mathtt{Dist}(\substring{x}_{\le m - 1}, \substring{y}_{\le n})\!+\!{}w_{\mathtt{I}} \\
                        \mathtt{Dist}(\substring{x}_{\le m}, \substring{y}_{\le n - 1})\!+\!{}w_{\mathtt{D}} \\
                        \mathtt{Dist}(\substring{x}_{\le m - 1}, \substring{y}_{\le n - 1})\!+\!{}w_{\mathtt{S}}(x_m, y_n)
                    \end{cases} %
    \end{aligned}%
\end{equation*}%
where $w_{\mathtt{I}}$, $w_{\mathtt{D}}$, and $w_{\mathtt{S}}(x_m, y_n)$ are the weights for insertion, deletion, and substitution, respectively.
For simplicity, the keep operation is treated as a special case of substitution.

\section{A Reinforcement Learning Perspective}
\label{sec:reinforcement_learning_perspective}
In this section, we provide a reinforcement learning perspective of our method.
We believe this perspective can help us understand the role of the two large language models in our method, and the relationship between our method and \texttt{TfPf}.

\subsection{KL-Regulated Reinforcement Learning}
Reinforcement learning (RL) has been widely used to improve the performance of LLMs by optimizing the following objective:
\begin{equation}
    \begin{aligned}
        \mathcal{L}(p_{\theta}) =\  & \mathbb{E}_{\boldsymbol{y}\sim p_{\theta}(\cdot\mid \boldsymbol{x})} \left[ r(\boldsymbol{x}, \boldsymbol{y}) \right]                   \\
                                    & - \beta D_{KL}\left(p_{\theta}\left(\cdot\mid \boldsymbol{x}\right) \Vert\ p_{\mathtt{ref}}\left(\cdot\mid \boldsymbol{x}\right)\right)
    \end{aligned}
\end{equation}
where $p_{\theta}$ is the model we want to optimize, $r$ is the reward function, $p_{\mathtt{ref}}$ is the reference model whose parameters are frozen, and $\beta$ is the KL-regularization coefficient that controls how much we want $p_{\theta}$ to be different from the reference model.
A larger $\beta$ means we want $p_{\theta}$ to be more similar to the reference model.

The optimal model $p^*$ of $p_{\theta}$ is unique and is given by:
\begin{equation}
    \begin{aligned}
        p^*(\boldsymbol{y}\mid \boldsymbol{x})\ \propto\ p_{\mathtt{ref}}(\boldsymbol{y}\mid \boldsymbol{x}) \exp\left(\frac{1}{\beta} r(\boldsymbol{x}, \boldsymbol{y})\right)%
    \end{aligned}%
    \label{eq:optimal_policy}
\end{equation}
This equation shows that we do not need to train the model to optimize the objective; instead, we can directly obtain the optimal policy by combining the reference model and the reward function during inference.

\subsection{\texttt{TfPf} as a Reward Model}
\label{subsec:tfpf_as_a_reward_model}
Intuitively, given an input $\substring{x}$ and two outputs $\substring{y}_{a}$ and $\substring{y}_{b}$, which output is better can be determined by the following two criteria:
\begin{asparaitem}[$\bullet$]
    \item \textbf{Fluency}: A better output $\substring{y}$ should be more fluent.
    \item \textbf{Faithfulness}: A better output $\substring{y}$ should only make the necessary changes to the input sentence $\substring{x}$. Between two sentences with the same fluency, the one with fewer changes is better.
\end{asparaitem}

Recall the score function of \texttt{TfPf} in Equation~\ref{eq:tfpf_score}.
The score function of \texttt{TfPf} is a combination of a large language model and a distance metric.
The large language model, acting as a pure language model, can be seen as a reward function measuring the fluency of the output.
On the other hand, the distance metric, acting as a reward function measuring the faithfulness of the output, penalizes outputs that change the input sentence too much.

Combining \texttt{TfPf} as a reward model with the RL framework, Equation~\ref{eq:optimal_policy} can be rewritten as:
\begin{equation}
    \begin{aligned}
        \log p^*(\substring{y}\mid \substring{x})\ \propto\  & \log p_{\mathtt{ref}}(\substring{y}\mid \substring{x})                                 \\
                                                             & + \frac{1}{\beta} \left(\begin{array}{c}\log p_{\mathtt{LLM}}(\substring{y}) \\
                                                                                               -                                       \\
                                                                                               \mathtt{Dist}(\substring{x}, \substring{y})\end{array}\right)
    \end{aligned}
\end{equation}

\paragraph{Link to the Original \texttt{TfPf} Paper}
If we set $\beta = 1$ and use a uniform distribution as the reference model $\log p_{\mathtt{ref}}(\substring{y}\mid \substring{x}) = C$, where $C$ is a constant, the above equation is equivalent to the score function of \texttt{TfPf} in Equation~\ref{eq:tfpf_score}.

\paragraph{Link to Our Method}
If we set $\beta = 1$ and use a prompt-based large language model as the reference model $\log p_{\mathtt{ref}}(\substring{y}\mid \substring{x}) = \log p_{\mathtt{LLM}}(\substring{y}\mid \mathtt{Prompt}(\substring{x}))$, the above equation becomes equivalent to our score function in Equation~\ref{eq:gcsc_score}.
By using a prompt-based large language model as the reference model instead of a uniform distribution, we obtain a more reasonable prior distribution, which enables our method to find better outputs $\substring{y}$ with a limited beam size.

\begin{figure*}[tb!]
    \centering%
    \begin{promptbox}{\texttt{Detailed Prompt}}{black}{white}
        \footnotesize{
            \textbf{\texttt{System Prompt:}}\\
            你是一个优秀的中文纠错模型，中文纠错模型即更正用户输入句子中的错误。\return

            \textbf{\texttt{User Prompt:}}\\
            你需要识别并纠正用户输入的句子中可能的错别字、多字、漏字并输出正确的句子，在修改的同时尽可能减少对原句子的改动(不新增、删除和修改标点符号)。只输出没有错误的句子，不要添加任何其他解释或说明。如果句子没有错误，就直接输出和输入相同的句子。\return

            输入：\jsonkey{\{INPUT: $\boldsymbol{x}$\}}\return
        }
        \tcblower
        \footnotesize
        \textbf{\texttt{System Prompt:}}\\
        You are an excellent Chinese error correction model, which means you correct errors in the sentences entered by the user.\return

        \footnotesize
        \textbf{\texttt{User Prompt:}}\\
        You need to identify and correct possible misspellings, redundant and missing characters in the user's input sentence, and output the correct sentence. While making corrections, try to minimize changes to the original sentence (without adding, deleting, or modifying punctuation). Only output the corrected sentence; do not add any further explanations or notes. If the sentence is error-free, output the exact same sentence as the input.\return

        \footnotesize
        Input: \jsonkey{\{INPUT: $\boldsymbol{x}$\}}\return

    \end{promptbox}
    \caption{
        A detailed prompt template.
    }
    \label{fig:detail_prompt}
\end{figure*}

\subsection{Proof of Equation \ref{eq:optimal_policy}}
\label{subsec:proof_of_optimal_policy}
First, we rewrite the objective function of RL as follows:
\begin{equation}
    \begin{aligned}
        \mathcal{L}(p_{\theta}) & = \mathbb{E}_{\boldsymbol{y}\sim p_{\theta}(\cdot\mid \boldsymbol{x})} \left[ r(\boldsymbol{x}, \boldsymbol{y}) \right]                                                                                                                                                            \\
                                & \quad - \beta D_{KL}\left(p_{\theta}\left(\cdot\mid \boldsymbol{x}\right) \Vert\ p_{\mathtt{ref}}\left(\cdot\mid \boldsymbol{x}\right)\right)                                                                                                                                      \\
                                & = \sum_{\boldsymbol{y}} p_{\theta}(\boldsymbol{y}\mid \boldsymbol{x}) \left(\begin{array}{c} r(\boldsymbol{x}, \boldsymbol{y}) \\
                                                                                                                      -                                     \\
                                                                                                                      \beta \log \frac{p_{\theta}(\boldsymbol{y}\mid \boldsymbol{x})}{p_{\mathtt{ref}}(\boldsymbol{y}\mid \boldsymbol{x})}\end{array}\right)
    \end{aligned}
\end{equation}

Then we can compute the gradient of the objective function with respect to the policy $p_{\theta}(\boldsymbol{y}\mid \boldsymbol{x})$:
\begin{equation}
    \frac{\partial \mathcal{L}(p_{\theta})}{\partial p_{\theta}(\boldsymbol{y}\mid \boldsymbol{x})} = r(\boldsymbol{x}, \boldsymbol{y}) - \beta \left(\log \frac{p_{\theta}(\boldsymbol{y}\mid \boldsymbol{x})}{p_{\mathtt{ref}}(\boldsymbol{y}\mid \boldsymbol{x})} + 1\right)
    \label{eq:gradient_of_objective_function}
\end{equation}

Now, we can find the optimal policy $p^*$ by setting the gradient of the objective function to zero:
\begin{equation}
    r(\boldsymbol{x}, \boldsymbol{y}) - \beta \left(\log \frac{p_{\theta}(\boldsymbol{y}\mid \boldsymbol{x})}{p_{\mathtt{ref}}(\boldsymbol{y}\mid \boldsymbol{x})} + 1\right) = 0
\end{equation}

By rearranging the equation and taking the exponential of both sides, we get:
\begin{equation}
    \begin{aligned}
        p^*(\boldsymbol{y}\mid \boldsymbol{x}) =\  & p_{\mathtt{ref}}(\boldsymbol{y}\mid \boldsymbol{x}) \exp\left(\frac{r(\boldsymbol{x}, \boldsymbol{y})}{\beta} - 1\right) \\
        \propto\                                   & p_{\mathtt{ref}}(\boldsymbol{y}\mid \boldsymbol{x}) \exp\left(\frac{r(\boldsymbol{x}, \boldsymbol{y})}{\beta}\right)
    \end{aligned}
\end{equation}
where the $-1$ term inside the exponential function can be safely ignored because it is a constant.

\begin{table*}[p!]
    \centering
    \setlength{\tabcolsep}{2.1pt}
    \begin{NiceTabular}{p{12em}ccccccc|cc|cc}
        \toprule
        \rowcolor[gray]{.9}
        \textbf{Datasets}                 & \Block[c]{1-7}{\textbf{Lemon}} &                 &              &                 &              &              &              & \Block[c]{1-2}{\textbf{CSCD-NS}} &               & \Block[c]{1-2}{\textbf{C2EC}} &               \\
        \textbf{Subsets}                  & \textit{Car}                   & \textit{Cot}    & \textit{Enc} & \textit{Gam}    & \textit{Mec} & \textit{New} & \textit{Nov} & \textit{dev}                     & \textit{Test} & \textit{dev}                  & \textit{Test} \\
        \midrule
        \textbf{Language}                 & \Block[c]{1-11}{Chinese}       &                 &              &                 &              &              &              &                                  &               &                               &               \\
        \midrule
        \textbf{All Sentences}            & 3,410                          & 1,026           & 3,434        & \phantom{0,}400 & 2,090        & 5,892        & 6,000        & 5,000                            & 5,000         & 1,995                         & 5,711         \\
        \textbf{Evaluation Sentences}     & 3,245                          & \phantom{0,}993 & 3,274        & \phantom{0,}393 & 1,942        & 5,887        & 6,000        & 5,000                            & 5,000         & 1,995                         & 5,711         \\
        \textbf{Erroneous Sentence Ratio} & 48.60                          & 44.41           & 48.59        & 37.66           & 46.60        & 49.96        & 50.23        & 46.28                            & 46.06         & 50.08                         & 49.92         \\
        \textbf{Average Length}           & 43.44                          & 40.11           & 39.95        & 32.81           & 39.18        & 25.15        & 36.24        & 57.45                            & 57.63         & 51.93                         & 41.88         \\
        \textbf{Average Error Character}  & \wz1.20                        & \wz1.10         & \wz1.12      & \wz1.10         & \wz1.13      & \wz1.11      & \wz1.13      & \wz1.10                          & \wz1.10       & \wz1.10                       & \wz1.16       \\
        \bottomrule
    \end{NiceTabular}
    \caption{
        The statistics of the datasets used in the experiments.
    }
    \label{tab:dataset_statistics}
\end{table*}

\section{Detailed Experiment Settings}
\label{sec:detailed_experiment_settings}
\subsection{Prompt Templates}
\label{sec:prompt_templates}
\paragraph{The Detailed Prompt for Our Method}
Figure~\ref{fig:detail_prompt} shows the detailed prompt for our method, which includes task-specific system prompts, format requirements, and additional notes.

\paragraph{Prompts for In-Context Learning Baselines}
The prompts for the \texttt{ICL} baseline are shown in Figure~\ref{fig:icl_detail_prompt}.
To ensure the performance of the \texttt{ICL} baseline, we use different prompts for the \texttt{ICL} baseline on conventional CSC and C2EC.
Specifically, when evaluating conventional CSC, we do not instruct the model to correct missing and redundant characters.

\subsection{An Approximate Implementation of Our Method}
\label{subsec:approximate_implementation_of_our_method}
To reduce computational cost and speed up the generation process, we make two approximations.
First, we approximate the incremental Levenshtein distance $\Delta_{\mathtt{Dist}_{\mathtt{L}}}(\substring{x}, a, b, \substring{t}_1\dots\substring{t}_{k-1}, \substring{t}_k)$ as $\mathtt{Dist}_{\mathtt{L}}(\substring{x}_{[a:b+1]}, \substring{t}_k)$.
Then, we reduce the search space by maintaining only one best end index $b$ for each $\substring{t}_1\dots\substring{t}_k$.
While these approximations may yield suboptimal results, they work well in practice.

\subsection{Dataset Statistics}
\label{subsec:dataset_statistics}
In this work, we use three datasets to evaluate the performance of our method.
All datasets used in this work are publicly available.
Specifically, the CSCD-NS dataset is publicly available under the MIT license, while the CCTC dataset is publicly available under the Apache 2.0 license.
The specifics of these datasets are listed in Table~\ref{tab:dataset_statistics}.
The \textbf{Evaluation Sentences} row in Table~\ref{tab:dataset_statistics} shows the number of sentences actually used for evaluation.
This is because sentences where the original and corrected versions differ in length are excluded when evaluating CSC models, as done in previous works \cite{liu-etal-2023-chinese,li-etal-2024-cllm,zhou-etal-2024-simple}.

\subsection{Evaluation Implementation and Settings}
\label{subsec:evaluation_implementation_and_settings}
We use the evaluation script from \citet{zhou-etal-2024-simple} to calculate the metrics.
This script adopts a Levenshtein distance algorithm to extract edit operations, allowing for comparisons between sentences of different lengths.
We also follow their settings by ignoring whitespaces and converting all full-width punctuation to half-width.

\subsection{Hardware Setup}
Experiments were conducted on a single NVIDIA A100 40GB GPU with the Intel Xeon Gold 6248R (3.00GHz) CPU.

\section{More Results}
\label{sec:detailed_results}
\subsection{Sentence-level Results}
\label{subsec:sentence_level_results}
\begin{table*}[p!]
    \centering
    \renewcommand{\arraystretch}{0.95}
    \setlength{\tabcolsep}{3.8pt}
    \scalebox{0.9}{
        \begin{NiceTabular}{lccccccccc;c;c|c}
            \toprule
            \rowcolor[gray]{.9}
                                                                                                                                             &                               &                                 & \Block[c]{1-9}{\textbf{\textsc{Conventional CSC}}} &                          &                          &                          &                          &                          &                          &                                  &                               & \textbf{\textsc{C2EC}} \\
            \Block[l]{2-1}{\textbf{Model}}                                                                                                   & \Block[c]{2-1}{\textbf{Size}} & \Block[c]{2-1}{\textbf{Method}} & \Block[c]{1-7}{\textbf{Lemon}}                     &                          &                          &                          &                          &                          &                          & \Block[c]{1-1}{\textbf{CSCD-NS}} & \Block[c]{2-1}{\textbf{Avg.}} & \textbf{C2EC}          \\
                                                                                                                                             &                               &                                 & \textit{Car}                                       & \textit{Cot}             & \textit{Enc}             & \textit{Gam}             & \textit{Mec}             & \textit{New}             & \textit{Nov}             & \textit{test}                    &                               & \textit{test}          \\
            \midrule
            \Block[c]{1-13}{\textit{Supervised Fine-tuning SoTAs} \cite{wu-etal-2023-rethinking,liu-etal-2024-rephrasing,hu-etal-2024-cscd}} &                               &                                 &                                                                                                                                                                                                                                                                                                                  \\
            \midrule
            \texttt{BERT}                                                                                                                    & \texttt{0.1B}                 & \texttt{SFT}                    & 52.3\rlap{$^\dagger$}\wz                           & 64.1\rlap{$^\dagger$}\wz & 45.5\rlap{$^\dagger$}\wz & 33.3\rlap{$^\dagger$}\wz & 50.9\rlap{$^\dagger$}\wz & 56.0\rlap{$^\dagger$}\wz & 36.0\rlap{$^\dagger$}\wz & 72.96\rlap{$^\ddagger$}          & --                            & --                     \\
            \texttt{SM-BERT}                                                                                                                 & \texttt{0.1B}                 & \texttt{SFT}                    & 52.0\rlap{$^\dagger$}\wz                           & 65.0\rlap{$^\dagger$}\wz & 44.6\rlap{$^\dagger$}\wz & 29.8\rlap{$^\dagger$}\wz & 49.3\rlap{$^\dagger$}\wz & 55.8\rlap{$^\dagger$}\wz & 37.8\rlap{$^\dagger$}\wz & \textbf{73.62}\rlap{$^\ddagger$} & --                            & --                     \\
            \texttt{ReLM}                                                                                                                    & \texttt{0.1B}                 & \texttt{SFT}                    & 53.1\rlap{$^\dagger$}\wz                           & 66.8\rlap{$^\dagger$}\wz & 49.2\rlap{$^\dagger$}\wz & 33.0\rlap{$^\dagger$}\wz & 54.0\rlap{$^\dagger$}\wz & 58.5\rlap{$^\dagger$}\wz & 37.8\rlap{$^\dagger$}\wz & --                               & --                            & --                     \\
            \Block[l]{2-1}{\texttt{Baichuan2}}                                                                                               & \texttt{7B}                   & \texttt{SFT}                    & --                                                 & --                       & --                       & --                       & --                       & --                       & --                       & 64.44\rlap{$^\ddagger$}          & --                            & --                     \\
                                                                                                                                             & \texttt{13B}                  & \texttt{SFT}                    & --                                                 & --                       & --                       & --                       & --                       & --                       & --                       & 66.10\rlap{$^\ddagger$}          & --                            & --                     \\
            \midrule
            \Block[c]{1-13}{\textit{BERT-based SFT Models cooperating with LLMs} \cite{liu-etal-2024-arm}}                                   &                               &                                 &                                                    &                          &                          &                          &                          &                          &                          &                                  &                               &                        \\
            \midrule
            \texttt{MDCSPell}$^\natural$                                                                                                     & \texttt{N/A}                  & \texttt{ARM}                    & 37.1\wz                                            & 52.7\wz                  & 35.2\wz                  & 15.3\wz                  & 33.0\wz                  & 36.4\wz                  & 15.6\wz                  & --                               &                               &                        \\
            \midrule
            \Block[c]{1-13}{\textit{Training-free Methods of LLMs}}                                                                          &                               &                                 &                                                    &                          &                          &                          &                          &                          &                          &                                  &                               &                        \\
            \midrule
            \texttt{GPT4o-mini}                                                                                                              & \texttt{N/A}                  & \texttt{ICL}                    & 29.43                                              & 30.85                    & 42.11                    & 26.67                    & 33.72                    & 33.12                    & 23.52                    & 38.78                            & 32.27                         & 29.09                  \\
            \texttt{GPT4o}                                                                                                                   & \texttt{N/A}                  & \texttt{ICL}                    & 52.91                                              & 59.98                    & 62.15                    & 39.11                    & 61.87                    & 61.82                    & 49.96                    & 63.89                            & 56.46                         & 47.66                  \\
            \texttt{DeepSeek\,V3}                                                                                                            & \texttt{671B}                 & \texttt{ICL}                    & 57.33                                              & \textbf{69.87}           & \textbf{66.07}           & \textbf{53.12}           & \textbf{69.33}           & 69.14                    & \textbf{57.43}           & 67.55                            & \textbf{63.73}                & 53.25                  \\
            \texttt{DeepSeek\,R1}                                                                                                            & \texttt{671B}                 & \texttt{ICL}                    & 54.52                                              & 63.43                    & 59.93                    & 49.27                    & 68.59                    & 67.82                    & 52.93                    & 62.39                            & 59.84                         & 45.58                  \\
            \hdashedline
            \Block[l]{4-1}{\texttt{Qwen2.5}}                                                                                                 & \Block[c]{4-1}{\texttt{7B}}   & \texttt{ICL}                    & 26.41                                              & 42.95                    & 37.97                    & 18.93                    & 38.27                    & 29.10                    & 21.85                    & 34.10                            & 31.20                         & 28.53                  \\
                                                                                                                                             &                               & \texttt{ICL-RR}                 & 38.93                                              & 56.46                    & 49.05                    & 32.40                    & 50.31                    & 44.54                    & 32.77                    & 54.40                            & 44.86                         & 40.24                  \\
                                                                                                                                             &                               & \texttt{TfPf}                   & 49.37                                              & 61.60                    & 48.48                    & 39.07                    & 55.69                    & 59.79                    & 39.09                    & 58.57                            & 51.46                         & 38.60                  \\
                                                                                                                                             &                               & \texttt{OUR}                    & 57.58                                              & 69.45                    & 61.56                    & 47.78                    & 66.59                    & 66.72                    & 47.91                    & 68.13                            & 60.72                         & 52.68                  \\
            \hdashedline
            \Block[l]{4-1}{\texttt{Qwen2.5}}                                                                                                 & \Block[c]{4-1}{\texttt{14B}}  & \texttt{ICL}                    & 35.91                                              & 52.31                    & 43.55                    & 32.86                    & 47.95                    & 40.70                    & 31.59                    & 41.54                            & 40.80                         & 31.73                  \\
                                                                                                                                             &                               & \texttt{ICL-RR}                 & 48.15                                              & 59.69                    & 53.99                    & 37.19                    & 55.60                    & 57.02                    & 42.38                    & 59.30                            & 51.67                         & 42.42                  \\
                                                                                                                                             &                               & \texttt{TfPf}                   & 50.25                                              & 59.92                    & 49.84                    & 32.88                    & 54.39                    & 60.94                    & 41.11                    & 57.95                            & 50.91                         & 37.84                  \\
                                                                                                                                             &                               & \texttt{OUR}                    & \textbf{59.83}                                     & 68.74                    & 64.37                    & 48.53                    & 66.54                    & \textbf{69.64}           & 49.65                    & 70.42                            & 62.22                         & \textbf{54.12}         \\
            \bottomrule
        \end{NiceTabular}
    }
    \caption{
        Sentence-level $F_1$ scores of our method and the baseline methods on conventional CSC datasets.
        $\dagger$ indicates the results are from models fine-tuned on 34M synthetic CSC data \cite{wu-etal-2023-rethinking,liu-etal-2024-rephrasing}.
        $\ddagger$ indicates the results are from models pre-trained on 2M synthetic data specifically designed for the CSCD-NS dataset before fine-tuning on the CSCD-NS training set \cite{hu-etal-2024-cscd}.
        $\natural$ The ARM method is based on the \texttt{GPT3.5 Turbo} model and a BERT-based model (MDCSPell), which is trained on 271k synthetic CSC data from \citet{wang-etal-2018-hybrid} and training data from the Sighan series datasets.
    }
    \label{tab:spelling_check_results.sf}
\end{table*}

The sentence-level $F_1$ results are shown in Table~\ref{tab:spelling_check_results.sf}.

The results show that our method outperforms three baselines on both conventional CSC and C2EC datasets.

Additionally, our method enables a 14B model to achieve performance comparable to the leading LLM, which has 671B parameters.

\subsection{A Fair Comparison with SFT Methods}
\label{app:supervised_fine_tuning}
\begin{table*}[t!]
    \centering
    \renewcommand{\arraystretch}{0.95}
    \scalebox{0.9}{
        \begin{NiceTabular}{lccccccccc;c;c}
            \toprule
            \Block[l]{2-1}{\textbf{Model}}   & \Block[c]{2-1}{\textbf{Size}} & \Block[c]{2-1}{\textbf{Method}} & \Block[c]{1-7}{\textbf{Lemon}} &                &                &                &                &                &                & \Block[c]{1-1}{\textbf{CSCD-NS}} & \Block[c]{2-1}{\textbf{Avg.}} \\
                                             &                               &                                 & \textit{Car}                   & \textit{Cot}   & \textit{Enc}   & \textit{Gam}   & \textit{Mec}   & \textit{New}   & \textit{Nov}   & \textit{test}                    &                               \\
            \midrule
            \texttt{SCOPE}                   & \texttt{0.1B}                 & \texttt{SFT-F}                  & 50.71                          & 54.89          & 45.23          & 24.74          & 44.44          & 48.72          & 33.17          & 71.70                            & 46.70                         \\
            \hdashedline
            \Block[l]{4-1}{\texttt{Qwen1.5}} & \Block[c]{4-1}{\texttt{7B}}   & \texttt{SFT-L}                  & 53.38                          & 56.55          & 54.44          & 37.33          & 59.21          & 58.96          & 39.12          & 68.66                            & 53.46                         \\
                                             &                               & \texttt{SFT-L}\rlap{$^\dagger$} & 53.87                          & 58.04          & 54.57          & 37.43          & 61.16          & 60.07          & 41.42          & \textbf{71.64}                   & 54.77                         \\
                                             &                               & \texttt{TfPf}                   & 53.88                          & 61.68          & 51.46          & 38.87          & 57.66          & 60.97          & 44.97          & 58.27                            & 53.47                         \\
                                             &                               & \texttt{OUR}                    & \textbf{61.57}                 & \textbf{69.10} & \textbf{63.34} & \textbf{48.50} & \textbf{65.34} & \textbf{68.89} & \textbf{50.27} & 67.25                            & \textbf{61.78}                \\
            \hdashedline
            \Block[l]{4-1}{\texttt{Qwen1.5}} & \Block[c]{4-1}{\texttt{14B}}  & \texttt{SFT-L}                  & 54.56                          & 56.82          & 53.44          & 32.59          & 58.89          & 63.32          & 40.58          & 72.63                            & 54.10                         \\
                                             &                               & \texttt{SFT-L}\rlap{$^\dagger$} & 57.54                          & 60.40          & 56.48          & 38.02          & 65.31          & 64.49          & 43.92          & \textbf{73.80}                   & 57.49                         \\
                                             &                               & \texttt{TfPf}                   & 52.61                          & 62.91          & 50.81          & 36.36          & 54.78          & 60.59          & 42.89          & 58.56                            & 52.44                         \\
                                             &                               & \texttt{OUR}                    & \textbf{62.88}                 & \textbf{70.31} & \textbf{66.24} & \textbf{46.59} & \textbf{66.67} & \textbf{70.15} & \textbf{52.69} & 71.53                            & \textbf{63.38}                \\
            \bottomrule
        \end{NiceTabular}
    }
    \caption{
        Fair comparison between our method and the supervised fine-tuning (SFT) methods.
        We adopt the SFT method from \citet{li-etal-2024-cllm}.
        \texttt{SFT-F} means the full parameter fine-tuning, while \texttt{SFT-L} means fine-tuning with LoRA.
        \texttt{SFT-L}$^\dagger$ means the C-LLM method from \citet{li-etal-2024-cllm}, that conducts the Character-level LoRA fine-tuning after the continuous pre-training.
    }
    \label{tab:main_results:sft}
\end{table*}

In main results, we compare our approach using the \texttt{Qwen2.5} series with the state-of-the-art methods that have been fine-tuned with supervision, as reported by \cite{li-etal-2024-cllm}.
It's important to note a potential discrepancy: their supervised fine-tuned methods were trained from the \texttt{Qwen1.5} series, whereas our method utilizes the \texttt{Qwen2.5} series.
To ensure a fair comparison, we also provide the results of our method on the \texttt{Qwen1.5} series in Table~\ref{tab:main_results:sft}.

When compared to the \texttt{TfPf} method, the \texttt{SFT} methods show superior performance on the in-domain dataset CSCD-NS.
However, they perform less effectively on the out-of-domain dataset Lemon, particularly with a 7B model.
This suggests that the \texttt{SFT} methods might overfit to the in-domain dataset CSCD-NS, limiting their generalization to the out-of-domain dataset Lemon.

Our method, which requires no training, significantly outperforms the \texttt{SFT} methods on the out-of-domain dataset Lemon.
Additionally, without any training, our method achieves performance on par with the \texttt{SFT} methods on the in-domain dataset CSCD-NS, with scores of 71.53 versus 73.80 on the 14B model.

\subsection{Additional Comparison on Sighan dataset}
\label{app:additional_comparison_on_sighan_dataset}
\begin{table*}[tb!]
    \setlength{\tabcolsep}{2.1pt}
    \renewcommand{\arraystretch}{0.95}
    \centering
    \scalebox{0.92}{
        \begin{NiceTabular}{lccc;ccc|ccc;ccc|ccc;ccc}
            \toprule
            \rowcolor[gray]{1.0}
            \Block[l]{3-1}{\textbf{Models}}                                                                  & \Block[c]{1-6}{\textbf{Sighan} \textit{13}} &               &                &                                     &               &                & \Block[c]{1-6}{\textbf{Sighan} \textit{14}} &               &                &                                     &               &                & \Block[c]{1-6}{\textbf{Sighan} \textit{15}} &               &                &                                     &               &                \\
                                                                                                             & \Block[c]{1-3}{\textbf{Detection}}          &               &                & \Block[c]{1-3}{\textbf{Correction}} &               &                & \Block[c]{1-3}{\textbf{Detection}}          &               &                & \Block[c]{1-3}{\textbf{Correction}} &               &                & \Block[c]{1-3}{\textbf{Detection}}          &               &                & \Block[c]{1-3}{\textbf{Correction}} &               &                \\
                                                                                                             & \textbf{P}                                  & \textbf{R}    & \textbf{F$_1$} & \textbf{P}                          & \textbf{R}    & \textbf{F$_1$} & \textbf{P}                                  & \textbf{R}    & \textbf{F$_1$} & \textbf{P}                          & \textbf{R}    & \textbf{F$_1$} & \textbf{P}                                  & \textbf{R}    & \textbf{F$_1$} & \textbf{P}                          & \textbf{R}    & \textbf{F$_1$} \\
            \midrule
            \rowcolor[gray]{.95}
            \Block[c]{1-19}{\textit{Supervised Fine-tuning SoTAs} (reported by \citet{qiao-etal-2024-disc})} &                                             &               &                &                                     &               &                &                                             &               &                &                                     &               &                &                                             &               &                &                                     &               &                \\
            \texttt{SpellGCN}                                                                                & 80.1                                        & 74.4          & 77.2           & 78.3                                & 72.7          & 75.4           & 65.1                                        & 69.5          & 67.2           & 63.1                                & 67.2          & 65.3           & 74.8                                        & 80.7          & 77.7           & 72.1                                & 77.7          & 75.9           \\
            \texttt{ReaLiSe}                                                                                 & 88.6                                        & 82.5          & 85.4           & 87.2                                & 81.2          & 84.1           & 67.8                                        & 71.5          & 69.6           & 66.3                                & 70.0          & 68.1           & 77.3                                        & 81.3          & 79.3           & 75.9                                & 79.9          & 77.8           \\
            \texttt{\quad+\,DISC}                                                                            & 88.9                                        & 82.2          & 85.4           & 87.6                                & 81.1          & 84.2           & 69.2                                        & 71.2          & 70.1           & 68.2                                & 70.2          & 69.2           & 78.3                                        & 81.2          & 79.7           & 77.0                                & 79.9          & 78.4           \\
            \texttt{SCOPE}                                                                                   & 87.5                                        & 83.0          & 85.2           & 86.5                                & 82.1          & 84.2           & 68.8                                        & 73.7          & 71.1           & 67.1                                & 71.2          & 69.5           & 80.5                                        & 85.4          & 82.9           & 78.7                                & 83.5          & 81.0           \\
            \texttt{\quad+\,DR-CSC}                                                                          & 88.5                                        & 83.7          & 86.0           & 87.7                                & 83.0          & 85.3           & \textbf{70.2}                               & 73.3          & 71.7           & 69.3                                & 72.3          & 70.7           & \textbf{82.9}                               & 84.8          & \textbf{83.8}  & \textbf{80.3}                       & 82.3          & 81.3           \\
            \texttt{\quad+\,DISC}                                                                            & 88.8                                        & 83.7          & 86.2           & 88.0                                & 83.0          & 85.4           & \textbf{70.2}                               & 73.5          & 71.8           & \textbf{69.3}                       & 72.5          & 70.9           & 81.7                                        & 84.8          & 83.2           & 80.2                                & 83.4          & \textbf{81.8}  \\
            \texttt{ReLM}                                                                                    & 86.4                                        & 83.7          & 85.0           & 85.0                                & 82.3          & 83.7           & 65.7                                        & 74.5          & 69.8           & 63.7                                & 72.3          & 67.7           & 78.3                                        & \textbf{85.6} & 81.8           & 76.8                                & \textbf{83.9} & 80.2           \\
            \texttt{\quad+\,DISC}                                                                            & \textbf{89.7}                               & \textbf{84.5} & \textbf{87.0}  & \textbf{88.4}                       & \textbf{83.3} & \textbf{85.8}  & 69.7                                        & \textbf{74.9} & \textbf{72.2}  & 68.6                                & \textbf{73.7} & \textbf{71.0}  & 80.8                                        & 84.3          & 82.5           & 79.8                                & 83.1          & 81.4           \\
            \midrule
            \rowcolor[gray]{.95}
            \Block[c]{1-19}{\textit{Training-free Methods}}                                                  &                                             &               &                &                                     &               &                &                                             &               &                &                                     &               &                &                                             &               &                &                                     &               &                \\
            \texttt{GPT3.5}                                                                                  & 61.6                                        & 29.2          & 39.7           & 57.1                                & 27.1          & 36.7           & 41.4                                        & 23.1          & 29.6           & 39.7                                & 22.1          & 28.4           & 39.4                                        & 46.4          & 42.6           & 32.7                                & 38.4          & 35.3           \\
            \texttt{GPT4}                                                                                    & 53.4                                        & 51.6          & 52.5           & 47.3                                & 45.7          & 46.5           & 38.1                                        & 52.3          & 44.1           & 32.8                                & 45.0          & 38.0           & 42.7                                        & 57.5          & 49.0           & 36.5                                & 49.2          & 41.9           \\
            \hdashedline
            \texttt{OUR Q2.5\,7B}                                                                            & 75.6                                        & 76.7          & 74.6           & 73.3                                & 74.4          & 72.3           & 51.1                                        & 47.2          & 55.6           & 48.9                                & 45.3          & 53.3           & 61.6                                        & 58.6          & 64.9           & 57.5                                & 54.8          & 60.6           \\
            \texttt{OUR Q2.5\,14B}                                                                           & 76.4                                        & 77.2          & 75.6           & 73.6                                & 74.3          & 72.8           & 51.9                                        & 47.7          & 56.9           & 49.5                                & 45.5          & 54.2           & 62.0                                        & 58.1          & 66.4           & 56.9                                & 53.4          & 61.0           \\
            \bottomrule
        \end{NiceTabular}
    }
    \caption{
        Sentence-level performance on the SIGHAN13, SIGHAN14 and SIGHAN15 test sets.
        We adopt the results of previous SOTAs from \citet{qiao-etal-2024-disc}.
        It is worth noting that all these models are trained on the SIGHAN series datasets.
        Apart from SpellGCN, all models apply post-processing on SIGHAN13, which removes all detected and corrected ``地'' and ``得'' from the model output before evaluation.
        `\texttt{Q2.5}' is short for \texttt{Qwen2.5} series models.
    }
    \label{tab:sighans_experiment}
\end{table*}

To provide a more comprehensive comparison, we also report the results of our method on the Sighan dataset, a CSC dataset, as well as some classical CSC models.

The results are shown in Table~\ref{tab:sighans_experiment}.
From the results, we can see that our method still lags behind classical CSC models, which are specifically trained on the Sighan dataset.

\subsection{Qualitative Analysis}
\label{subsec:qualitative_analysis}
\begin{table*}[p!]
    \setlength{\tabcolsep}{2.5pt}
    \renewcommand{\arraystretch}{1.1}
    \centering
    {
        \scalebox{0.9}{
            \begin{NiceTabular}{p{2.5cm}l}
                \toprule
                \rowcolor[gray]{.95}
                Input           & 醋酸\wrong{圈}奈德尿素乳膏是一种复方的外用软膏,主要含有醋酸曲安奈德和尿素。       \\
                Reference       & 醋酸\correct{曲安}奈德尿素乳膏是一种复方的外用软膏,主要含有醋酸曲安奈德和尿素。    \\
                \midrule
                \texttt{ICL}    & 醋酸\correct{曲安}奈德尿素乳膏是一种复方的外用软膏,主要含有醋酸曲安奈德和尿素。    \\
                \texttt{ICL-RR} & 醋酸\correct{曲安}奈德尿素乳膏是一种复方的外用软膏,主要含有醋酸曲安奈德和尿素。    \\
                \texttt{TfPf}   & 醋酸\wrong{甲}奈德尿素乳膏是一种复方的外用软膏,主要含有醋酸曲安奈德和尿素。       \\
                \hdashedline
                \texttt{OUR}    & 醋酸\correct{曲安}奈德尿素乳膏是一种复方的外用软膏,主要含有醋酸曲安奈德和尿素。    \\
                \midrule
                \midrule
                \rowcolor[gray]{.95}
                Input           & 无效者不需任何理由和解释凭收据10日内免费\wrong{凋}方或按余药退款。           \\
                Reference       & 无效者不需任何理由和解释凭收据10日内免费\correct{调}方或按余药退款。         \\
                \midrule
                \texttt{ICL}    & 无效者不需任何理由和解释凭收据10日内免费\correct{调}方或按余药退款。         \\
                \texttt{ICL-RR} & 无效者不需任何理由和解释凭收据10日内免费\correct{调}方或按余药退款。         \\
                \texttt{TfPf}   & 无效者不需任何理由和解释凭收据10日内免费\correct{调}\wrong{换}或按余药退款。 \\
                \hdashedline
                \texttt{OUR}    & 无效者不需任何理由和解释凭收据10日内免费\correct{调}方或按余药退款。         \\
                \midrule
                \midrule
                \rowcolor[gray]{.95}
                Input           & 当由于上述原因而\wrong{试}乙方无法履行保证时,\wrong{以}方不承担责任。      \\
                Reference       & 当由于上述原因而\correct{使}乙方无法履行保证时,\correct{乙}方不承担责任。  \\
                \midrule
                \texttt{ICL}    & 当由于上述原因而\correct{使}乙方无法履行保证时,\wrong{甲}方不承担责任。    \\
                \texttt{ICL-RR} & 当由于上述原因而\correct{使}乙方无法履行保证时,\wrong{甲}方不承担责任。    \\
                \texttt{TfPf}   & 当由于上述原因而\correct{使}乙方无法履行保证时,\correct{乙}方不承担责任。  \\
                \hdashedline
                \texttt{OUR}    & 当由于上述原因而\correct{使}乙方无法履行保证时,\correct{乙}方不承担责任。  \\
                \midrule
                \midrule
                \rowcolor[gray]{.95}
                Input           & 当月20座大城市房价同比增长4.9\%,创下2012年10月来最小增速              \\
                Reference       & 当月20座大城市房价同比增长4.9\%,创下2012年10月来最小增速。             \\
                \midrule
                \texttt{ICL}    & 当月20座大城市房价同比增长4.9\%,创下2012年10月来最小增速。             \\
                \texttt{ICL-RR} & 当月20座大城市房价同比增长4.9\%,创下2012年10月\wrong{以}来最小增速。    \\
                \texttt{TfPf}   & 当月20座大城市房价同比增长4.9\%,创下2012年10月来最小增速。             \\
                \hdashedline
                \texttt{OUR}    & 当月20座大城市房价同比增长4.9\%,创下2012年10月\wrong{以}来最小增速。    \\
                \bottomrule
            \end{NiceTabular}
        }
    }
    \caption{
        Qualitative examples of our approach and the baselines using the \texttt{Qwen2.5\;14B} model.
        Corrections marked in ``\correct{Blue}'' are correct, while those in ``\wrong{Red}'' are incorrect.
    }
    \label{tab:qualitative_examples}
\end{table*}

Table~\ref{tab:qualitative_examples} shows four examples illustrating the effectiveness of our method.

In the first example, characters ``曲安'' (\textit{qū ān}) were incorrectly typed as ``圈'' (\textit{quān}).
This simple error highlights the limitation of the Hamming distance in \texttt{TfPf}, which led to an incorrect correction to ``甲'' (\textit{jiǎ}) instead of the correct insertion of a character.

The second example is the character ``调'' (\textit{diào}) in ``调方'' (\textit{prescription adjustment}), which was mistakenly entered as ``凋'' (\textit{diāo}).
Correcting this error requires subsequent contextual understanding, which pure LLM probabilities in \texttt{TfPf} failed to achieve, resulting in an incorrect high-frequency substitution to ``调换'' (\textit{exchange, diào huàn}).
This issue was mitigated by incorporating prompt-based LLM probabilities.

In the third example, the character ``乙'' (\textit{A, yǐ}) was mistyped as ``以'' (\textit{yǐ}).
The \texttt{ICL} baseline overlooked phonetic similarities, incorrectly changing it to ``甲'' (\textit{A, jiǎ}).
The \texttt{ICL-RR} baseline also failed to correct this error as the correction ``以''$\rightarrow$``乙'' was not among the top-K candidates.

The fourth example is a negative case where our method unnecessarily inserted an ``以'' into an already correct sentence.
While this change might make the sentence slightly smoother, it is an over-correction since the original sentence was error-free.

\subsection{Incorrect Thinking of Reasoning Model May Lead to Wrong Corrections}
\label{subsec:incorrect_thinking_may_lead_to_wrong_corrections}
\begin{figure*}[ph!]
    \newtcolorbox{scopebox}[3]{%
        left=0pt,
        right=0pt,
        top=0pt,
        bottom=0pt,
        boxsep=3pt,
        middle=0pt,
        colback={#3},
        colframe={#2},
        title={#1},
    }
    \centering
    \footnotesize
    \begin{scopebox}{Example 1: \textit{Correct Thinking (with a little flaw)}}{black}{white}
        \begin{scopebox}{\texttt{Input}}{black!45}{white}
            松原市开展清\wrong{埋}整顿和\wrong{视}范房地产市场大检查
        \end{scopebox}
        \begin{scopebox}{\texttt{Reference}}{json_blue!60}{white}
            松原市开展清\correct{理}整顿和\correct{规}范房地产市场大检查
            \tcblower
            \textit{The city of Songyuan is conducting a major inspection to clean up, regulate, and standardize the real estate market.}
        \end{scopebox}
        \begin{scopebox}{\texttt{DeepSeek\,V3 Output}}{json_blue!80}{white}
            松原市开展清\correct{理}整顿和\correct{规}范房地产市场大检查
        \end{scopebox}
        \begin{scopebox}{\texttt{DeepSeek\,R1 Think}}{thinking_yellow}{white}
            \qquad{}好的，我现在需要处理用户输入的句子，找出可能的错别字并进行纠正，同时尽量减少对原句的改动，不新增、删除或修改标点符号。首先，我要仔细阅读用户提供的例子，了解他们期望的处理方式。\return

            \qquad{}用户给的例子中，比如第一个输入中的“班马王子”被纠正为“斑马王子”，这里“班”和“斑”是同音字，但“斑马”是正确的写法。另一个例子是“自然延生”改为“自然诞生”，这里“延”和“诞”字形相近，但“诞生”是正确的。还有“登录”改为“登陆”，虽然发音相同，但“登陆”在这里是正确的动词。这些例子显示，需要关注同音字、形近字以及特定语境下的正确用词。\return

            \qquad{}现在来看当前的输入句子：“松原市开展清埋整顿和视范房地产市场大检查”。我需要逐词分析可能的错别字。\return

            \qquad{}首先，“清埋”这个词是否正确？正确的应该是“清理”吗？\wrong{因为“埋”和“理”发音相近}，但“清理整顿”是常见的搭配，比如清理市场、整顿秩序。所以这里“清埋”可能是“清理”的错别字。\return

            \qquad{}接下来是“视范”，正确的应该是“规范”吗？“视”和“规”在字形上有点相似，但发音不同。“规范房地产市场”是合理的表达，而“视范”这个词组不太常见，可能是“规范”的笔误或形近错误。\return

            \qquad{}因此，纠正后的句子应该是：“松原市开展清理整顿和规范房地产市场大检查”。这样改动后的句子只替换了两个字，符合用户要求的最小改动原则，并且没有影响标点符号。\return

            \qquad{}再检查一遍是否有其他可能的错误。比如“整顿”是否正确？是的，没有问题。“房地产市场”也是正确的。标点符号也没有需要改动的地方。所以最终的输出应该是纠正后的这个句子。\return

            \tcblower
            \qquad{}\textit{Okay, I now need to process the user's input sentence, identify possible typos, and correct them while minimizing changes to the original sentence, without adding, deleting, or modifying punctuation. First, I need to carefully read the examples provided by the user to understand their expected handling method.}\return

            \qquad{}\textit{In the examples given by the user, for instance, the first input "班马王子" was corrected to "斑马王子", where "班" and "斑" are homophones, but "斑马" is the correct spelling. Another example is "自然延生" changed to "自然诞生", where "延" and "诞" have similar shapes, but "诞生" is correct. Also, "登录" changed to "登陆", although they sound the same, "登陆" is the correct verb here. These examples show the need to pay attention to homophones, similar-looking characters, and the correct word usage in specific contexts.}\return

            \qquad{}\textit{Now, let's look at the current input sentence: "松原市开展清埋整顿和视范房地产市场大检查". I need to analyze each word for possible typos.}\return

            \qquad{}\textit{First, is the word "清埋" correct? Should it be "清理"? \wrong{Because "埋" and "理" sound similar}, but "清理整顿" is a common phrase, like cleaning up the market and rectifying order. So "清埋" might be a typo for "清理".}\return

            \qquad{}\textit{Next is "视范", should it be "规范"? "视" and "规" look somewhat similar, but sound different. "规范房地产市场" is a reasonable expression, while "视范" is not common and might be a typo or similar-looking error for "规范".}\return

            \qquad{}\textit{Therefore, the corrected sentence should be: "松原市开展清理整顿和规范房地产市场大检查". This revised sentence only replaces two characters, adhering to the user's principle of minimal changes, and does not affect punctuation.}\return

            \qquad{}\textit{Check again for any other possible errors. For example, is "整顿" correct? Yes, no problem. "房地产市场" is also correct. There is no need to change punctuation. So the final output should be this corrected sentence.}\return

        \end{scopebox}
        \begin{scopebox}{\texttt{DeepSeek\,R1 Output}}{json_blue!80}{white}
            松原市开展清\correct{理}整顿和\correct{规}范房地产市场大检查
        \end{scopebox}
    \end{scopebox}
    \caption{
        An example of {Correct Thinking}
    }
    \label{fig:thinking_example_1}
\end{figure*}

\begin{figure*}[ph!]
    \newtcolorbox{scopebox}[3]{%
        left=0pt,
        right=0pt,
        top=0pt,
        bottom=0pt,
        boxsep=3pt,
        middle=0pt,
        colback={#3},
        colframe={#2},
        title={#1},
    }
    \centering
    \footnotesize
    \begin{scopebox}{Example 2: \textit{Correct Thinking}}{black}{white}
        \begin{scopebox}{\texttt{Input}}{black!45}{white}
            目前\wrong{上}暂无任何一种治疗方式可保证高热惊厥不复发。
        \end{scopebox}
        \begin{scopebox}{\texttt{Reference}}{json_blue!60}{white}
            目前暂无任何一种治疗方式可保证高热惊厥不复发。
            \tcblower
            \textit{Currently, there is no treatment available that can guarantee the prevention of recurrent febrile seizures.}
        \end{scopebox}
        \begin{scopebox}{\texttt{DeepSeek\,V3 Output}}{figure_red!80}{white}
            目前\wrong{尚}暂无任何一种治疗方式可保证高热惊厥不复发。
        \end{scopebox}
        \begin{scopebox}{\texttt{DeepSeek\,R1 Think}}{thinking_yellow}{white}
            \qquad{}好的，我现在需要处理用户的输入，找出可能的错别字、多字或漏字，并尽可能少地修改原句。首先，用户提供的例子显示，他们希望保持标点符号不变，只调整文字部分。\return

            \qquad{}现在来看当前的输入句子：“目前上暂无任何一种治疗方式可保证高热惊厥不复发。” 我需要仔细分析每个部分是否有错误。\return

            \qquad{}首先，“目前上暂无”这里可能有问题。“目前上”听起来不太自然，正确的表达应该是“目前”或者“目前临床上”。但原句中的“上”是否多余呢？比如，正确的说法可能是“目前暂无”，而“上”可能是多出来的字。例如，用户可能想表达“目前暂无”，但误加了“上”字。因此，“目前上”应该改为“目前”，去掉“上”字。\return

            \qquad{}接下来检查其他部分：“暂无任何一种治疗方式可保证高热惊厥不复发。” 这里的“暂无”是否正确？“暂无”表示暂时没有，符合语境。其他词语如“治疗方式”、“保证”、“复发”都没有明显错误。标点符号也没有问题，句尾是句号，符合要求。\return

            \qquad{}所以，修改后的句子应该是：“目前暂无任何一种治疗方式可保证高热惊厥不复发。” 去除了多余的“上”字，使句子更通顺，同时没有改动其他部分，符合用户的要求。\return

            \tcblower
            \qquad{}\textit{Alright, I now need to process the user's input, identify possible typos, extra words, or missing words, and make minimal changes to the original sentence. First, the examples provided by the user show that they want to keep the punctuation unchanged and only adjust the text part.}\return

            \qquad{}\textit{Now, let's look at the current input sentence: "目前上暂无任何一种治疗方式可保证高热惊厥不复发." I need to carefully analyze each part for errors.}\return

            \qquad{}\textit{First, "目前上暂无" might have an issue. "目前上" doesn't sound natural; the correct expression should be "目前" or "目前临床上". But is the "上" in the original sentence redundant? For example, the correct phrase might be "目前暂无", and "上" might be an extra word. For example, the user might have intended to say "目前暂无" but mistakenly added "上". Therefore, "目前上" should be changed to "目前", removing the "上".}\return

            \qquad{}\textit{Next, check the other parts: "暂无任何一种治疗方式可保证高热惊厥不复发." Is "暂无" correct here? "暂无" means temporarily unavailable, which fits the context. Other words like "治疗方式", "保证", "复发" have no obvious errors. The punctuation is also correct, with a period at the end, meeting the requirements.}\return

            \qquad{}\textit{So, the revised sentence should be: "目前暂无任何一种治疗方式可保证高热惊厥不复发." The extra "上" is removed, making the sentence smoother, while no other parts are changed, meeting the user's requirements.}\return
        \end{scopebox}
        \begin{scopebox}{\texttt{DeepSeek\,R1 Output}}{json_blue!80}{white}
            目前暂无任何一种治疗方式可保证高热惊厥不复发。
        \end{scopebox}
    \end{scopebox}
    \caption{
        Another example of {Correct Thinking}
    }
    \label{fig:thinking_example_2}
\end{figure*}

\begin{figure*}[ph!]
    \newtcolorbox{scopebox}[3]{%
        left=0pt,
        right=0pt,
        top=0pt,
        bottom=0pt,
        boxsep=3pt,
        middle=0pt,
        colback={#3},
        colframe={#2},
        title={#1},
    }
    \centering
    \footnotesize
    \begin{scopebox}{Example 2: \textit{Overthinking}}{black}{white}
        \begin{scopebox}{\texttt{Input}}{black!45}{white}
            新抚区已清理违法刀匾近70处将扩大整治范围发现一处拆除一处
        \end{scopebox}
        \begin{scopebox}{\texttt{Reference}}{json_blue!60}{white}
            新抚区已清理违法刀匾近70处将扩大整治范围发现一处拆除一处
            \tcblower
            \textit{Xinfu District has removed nearly 70 illegal protruding vertical signboards and will expand the rectification area - removing them as they are discovered.}
        \end{scopebox}
        \begin{scopebox}{\texttt{DeepSeek\,V3 Output}}{json_blue!80}{white}
            新抚区已清理违法刀匾近70处将扩大整治范围发现一处拆除一处
        \end{scopebox}
        \begin{scopebox}{\texttt{DeepSeek\,R1 Think}}{thinking_yellow}{white}
            \qquad{}好的，我现在需要处理用户的输入，找出可能的错别字并进行纠正，同时尽量少改动原句，不增删或修改标点。首先，我需要仔细阅读用户提供的例子，理解他们的处理方式和规则。\return

            \qquad{}用户的输入是：“新抚区已清理违法刀匾近70处将扩大整治范围发现一处拆除一处”。我要逐词分析，看看有没有可能的错别字。首先，“刀匾”这个词听起来不太对，可能是一个错误。通常我们会说“刀牌”或者“招牌”，但“刀匾”似乎不太常见。可能用户是想表达“刀牌”，而“匾”可能是一个错别字，正确的应该是“刀牌”或者“招牌”。比如，违法设置的广告牌、招牌等，所以这里“刀匾”可能是“刀牌”的误写，或者“招牌”的错误。但“刀牌”也不太常见，可能是“招牌”中的“招”被误写为“刀”。因为“招”和“刀”在字形上有点相似，尤其是在手写或快速输入时容易出错。所以可能的正确词是“招牌”，即“违法招牌”。\return

            \qquad{}接下来，验证这个推测是否符合语境。原文在说清理违法设置的物品，通常这类整治行动针对的是违规广告牌、招牌等，所以“招牌”更符合语境。因此，“刀匾”应纠正为“招牌”。这样修改后，句子变为：“新抚区已清理违法招牌近70处将扩大整治范围发现一处拆除一处”。检查其他部分是否有错误，比如“近70处将扩大”这里的“将”是否正确，是的，没有问题。其他部分也没有明显的错别字，所以最终的纠正应该是将“刀匾”改为“招牌”。\return

            \qquad{}另外，需要注意标点符号是否被改动，原句中没有标点，所以输出也应该保持无标点，但根据例子中的情况，用户可能允许保持原标点结构。不过在这个输入中，原句没有使用标点，所以输出也不需要添加。但可能用户希望保持原样，所以不需要添加标点。因此，最终输出应为：“新抚区已清理违法招牌近70处将扩大整治范围发现一处拆除一处”。\return
            \tcblower
            \qquad{}\textit{Alright, I now need to process the user's input, identify possible typos, and make corrections while minimizing changes to the original sentence, without adding, deleting, or modifying punctuation. First, I need to carefully read the examples provided by the user to understand their processing methods and rules.}\return

            \qquad{}\textit{The user's input is: "新抚区已清理违法刀匾近70处将扩大整治范围发现一处拆除一处". I need to analyze word by word to see if there are any possible typos. First, the word "刀匾" doesn't sound quite right and might be an error. We usually say "刀牌" or "招牌", but "刀匾" seems uncommon. The user might mean "刀牌", and "匾" might be a typo, with the correct word being "刀牌" or "招牌". For example, illegal advertising signs, signboards, etc., so here "刀匾" might be a miswriting of "刀牌" or an error for "招牌". However, "刀牌" is also uncommon, and it might be that "招" in "招牌" was mistakenly written as "刀". Since "招" and "刀" are somewhat similar in shape, especially in handwriting or fast input, they are easy to confuse. Therefore, the correct word might be "招牌", meaning "illegal signboards".}\return

            \qquad{}\textit{Next, verify if this hypothesis fits the context. The original text talks about clearing illegally set items, and such rectification actions usually target illegal advertising signs, signboards, etc., so "招牌" fits the context better. Therefore, "刀匾" should be corrected to "招牌". After this modification, the sentence becomes: "新抚区已清理违法招牌近70处将扩大整治范围发现一处拆除一处". Check if there are errors in other parts, such as whether "将" in "近70处将扩大" is correct. Yes, there is no problem. There are no obvious typos in other parts, so the final correction should be changing "刀匾" to "招牌".}\return

            \qquad{}\textit{Additionally, pay attention to whether punctuation has been altered. The original sentence has no punctuation, so the output should also remain without punctuation. However, based on the examples, the user might allow maintaining the original punctuation structure. But in this input, the original sentence does not use punctuation, so the output does not need to add any. But, the user might want to keep it as is, so no punctuation needs to be added. Therefore, the final output should be: "新抚区已清理违法招牌近70处将扩大整治范围发现一处拆除一处".}\return
        \end{scopebox}
        \begin{scopebox}{\texttt{DeepSeek\,R1 Output}}{figure_red!80}{white}
            新抚区已清理违法\wrong{招牌}近70处将扩大整治范围发现一处拆除一处
        \end{scopebox}
    \end{scopebox}
    \caption{
        An example of {Incorrect Thinking}
    }
    \label{fig:thinking_example_3}
\end{figure*}

\begin{figure*}[ph!]
    \newtcolorbox{scopebox}[3]{%
        left=0pt,
        right=0pt,
        top=0pt,
        bottom=0pt,
        boxsep=3pt,
        middle=0pt,
        colback={#3},
        colframe={#2},
        title={#1},
    }
    \centering
    \footnotesize
    \begin{scopebox}{Example 3: \textit{Oversight}}{black}{white}
        \begin{scopebox}{\texttt{Input}}{black!45}{white}
            行窃\wrong{是}触电身亡,小偷家属索赔,赔还是不赔
        \end{scopebox}
        \begin{scopebox}{\texttt{Reference}}{json_blue!60}{white}
            行窃\correct{时}触电身亡,小偷家属索赔,赔还是不赔
            \tcblower
            Died of electrocution while stealing, the thief's family claims compensation—compensate or not.
        \end{scopebox}
        \begin{scopebox}{\texttt{DeepSeek\,V3 Output}}{json_blue!80}{white}
            行窃\correct{时}触电身亡,小偷家属索赔,赔还是不赔
        \end{scopebox}
        \begin{scopebox}{\texttt{DeepSeek\,R1 Think}}{thinking_yellow}{white}
            \qquad{}好，我现在需要处理用户输入的句子，找出可能的错别字并进行纠正，同时尽量少改动原句。首先，我需要仔细阅读输入的句子：\return

            \qquad{}输入: 行窃是触电身亡,小偷家属索赔,赔还是不赔\return

            \qquad{}首先，逐词检查有没有明显的错别字。第一个部分“行窃是触电身亡”，看起来没问题，“行窃”指的是偷窃行为，“触电身亡”正确。接着是“小偷家属索赔”，这里“家属”正确，没有错别字。最后是“赔还是不赔”，这里“赔”字重复使用，但结构正确，没有问题。\return

            \qquad{}再检查标点符号是否正确。用户使用了逗号分隔各部分，没有使用错误，比如句号或分号，所以标点符号不需要改动。\return

            \qquad{}接下来，考虑是否有同音字或形近字错误。例如，“触电”是否应为“触电”？是的，正确。“身亡”正确。其他词汇如“行窃”、“家属”、“索赔”都正确。\return

            \qquad{}有没有可能被忽略的错误？比如“行窃”是否应为“行窃”？不，是正确的。“触电”是否有其他写法？比如“电击”，但原句用“触电”更准确，指接触电流导致死亡，正确。\return

            \qquad{}所以整个句子看起来没有错别字，结构和用词都正确。因此，输出应该与输入相同，不需要修改。\return

            \tcblower
            \qquad{}\textit{Ok, I now need to process the user's input sentence, identify possible typos, and make corrections while minimizing changes to the original sentence. First, I need to carefully read the input sentence:}\return

            \qquad{}\textit{Input: 行窃是触电身亡,小偷家属索赔,赔还是不赔}\return

            \qquad{}\textit{First, check word by word for any obvious typos. The first part "行窃是触电身亡" seems fine, "行窃" refers to the act of stealing, "触电身亡" is correct. Next is "小偷家属索赔", here "家属" is correct, no typos. Finally, "赔还是不赔", the word "赔" is repeated, but the structure is correct, no issues.}\return

            \qquad{}\textit{Then check if the punctuation is correct. The user used commas to separate each part, no errors like periods or semicolons, so the punctuation does not need to be changed.}\return

            \qquad{}\textit{Next, consider if there are homophones or similar-looking character errors. For example, should "触电" be "触电"? Yes, it's correct. "身亡" is correct. Other words like "行窃", "家属", "索赔" are all correct.}\return

            \qquad{}\textit{Are there any errors that might have been overlooked? For example, should "行窃" be "行窃"? No, it's correct. Is there another way to write "触电"? Like "电击", but the original sentence uses "触电" more accurately, referring to death by electric shock, correct.}\return

            \qquad{}\textit{So the whole sentence seems to have no typos, the structure and wording are correct. Therefore, the output should be the same as the input, no modifications needed.}\return
        \end{scopebox}
        \begin{scopebox}{\texttt{DeepSeek\,R1 Output}}{figure_red!80}{white}
            行窃\wrong{是}触电身亡,小偷家属索赔,赔还是不赔
        \end{scopebox}
    \end{scopebox}
    \caption{
        Another example of {Incorrect Thinking}
    }
    \label{fig:thinking_example_4}
\end{figure*}

The reasoning model \texttt{Deepseek\,R1} \cite{deepseek-r1-2025} shows very impressive performance on tasks like math, code, and science.
However, as discussed in the main results, \texttt{Deepseek\,R1} achieves lower performance on CSC tasks than its non-reasoning variant \texttt{Deepseek\,V3}.
Although \texttt{Deepseek\,R1} achieves higher recall than \texttt{Deepseek\,V3}, it introduces too many over-corrections.

After analyzing the reasoning process of \texttt{Deepseek\,R1}, we find that incorrect thinking may lead to wrong corrections, resulting in lower performance than \texttt{Deepseek\,V3}.

\Cref{fig:thinking_example_1,fig:thinking_example_2,fig:thinking_example_3,fig:thinking_example_4} show thinking examples from \texttt{Deepseek\,R1}.

The first two examples are correct thinking examples, showing that \texttt{Deepseek\,R1} can make reasonable thinking and correctly fix errors based on it.
Minor flaws in thinking may not affect the correctness of the correction.
For example, in the first example, \texttt{Deepseek\,R1} considers ``埋'' (\textit{mái}) and ``理'' (\textit{lǐ}) to be similar-sounding, but in fact they are look-alikes rather than sound-alikes.
However, this does not affect the correction result.

In the third example, ``刀匾'' (\textit{\ul{vertical} signboards}, \textit{dāo biǎn}) is an uncommon but correct word.
However, \texttt{Deepseek\,R1} over-thinks about it and incorrectly changes it to ``招牌'' (\textit{signboard}, \textit{zhāo pái}), a more common word with a similar but slightly different meaning.

In the fourth example, \texttt{Deepseek\,R1} analyzes all words in the input sentence but fails to consider whether the ``是'' in the input sentence is correct or not.
This oversight leads to an under-correction.

Nevertheless, we believe that the potential of the reasoning model \texttt{Deepseek\,R1} is still huge.
We plan to further investigate the use of reasoning models in CSC and C2EC tasks in the future.

\section{More Discussions}
\label{app:more_discussion}

\subsection{Impact of Beam Size}
\label{app:beam_size}
\begin{figure}[tb!]
    \centering
    \captionsetup[subfigure]{skip=-1pt, margin=5pt}
    \begin{tikzpicture}[
            legend/.style={
                    fill=white,
                    font=\footnotesize,
                    inner sep=2pt,
                    minimum width=1.0cm,
                    text opacity=1.0,
                    fill opacity=1.0,
                },
            diff label/.style={
                    font=\scriptsize,
                    inner sep=0.5pt,
                    outer sep=1.5pt,
                    fill=white,
                    fill opacity=0.9,
                    text opacity=1.0,
                    rounded corners=1pt,
                },
            trim left
        ]
        \centering
        \begin{groupplot}[
                group style={
                        group size=1 by 2,
                        x descriptions at=edge bottom,
                        horizontal sep=0.7cm,
                        vertical sep=0.1cm,
                    },
                width=1.0\linewidth,
                height=0.7\linewidth,
                xlabel={Beam Size},
                xmin=1.5,
                xmax=12.5,
                xmajorgrids=true,
                xtick={2,4,6,8,10,12},
                xticklabels={2,4,6,8,10,12},
                /tikz/font=\footnotesize,
                ylabel shift=-4pt,
                xlabel shift=-4pt,
                yticklabel shift=-2pt,
                xticklabel shift=-1pt,
                legend style={
                        at={(0.5,1.05)},
                        anchor=south,
                        font=\footnotesize,
                    },
                legend columns=4,
            ]
            \nextgroupplot[ymin=28,ymax=80]
            \addplot+ [mark=triangle*, draw=black, thick,
                densely dashed,
                mark size=2.5pt,
                mark options={fill=white, fill opacity=1.0, solid},
                opacity=1.0,
            ] table [row sep=\\] {
                    x	y\\
                    2	31.82\\
                    4	31.90\\
                    6	31.79\\
                    8	31.65\\
                    10	31.74\\
                    12	31.77\\
                };
            \addplot+ [mark=*, draw=black, thick,
                mark size=1.25pt,
                mark options={fill=white, fill opacity=1.0, solid},
                opacity=1.0,
            ] table [row sep=\\] {
                    x	y\\
                    2	40.82\\
                    4	48.17\\
                    6	51.19\\
                    8	52.59\\
                    10	53.86\\
                    12	54.54\\
                };
            \addplot+ [mark=square*, draw=figure_red, thick,
                dashed,
                mark size=1.25pt,
                mark options={fill=white, fill opacity=1.0, solid},
                opacity=1.0,
            ] table [row sep=\\] {
                    x	y\\
                    2	53.42\\
                    4	59.26\\
                    6	60.79\\
                    8	61.11\\
                    10	61.24\\
                    12	61.22\\
                };
            \addplot+ [mark=*, draw=figure_blue, thick,
                mark size=1.25pt,
                mark options={fill=figure_blue, fill opacity=1.0, solid},
                opacity=1.0,
            ] table [row sep=\\] {
                    x	y\\
                    2	68.18\\
                    4	70.56\\
                    6	70.52\\
                    8	70.57\\
                    10	70.30\\
                    12	70.06\\
                };
            \legend{\texttt{ICL}, \texttt{ICL-RR}, \texttt{TfPf},\texttt{OUR}}
            \nextgroupplot[ymin=25,ymax=65]
            \addplot+ [mark=triangle*, draw=black, thick,
                densely dashed,
                mark size=2.5pt,
                mark options={fill=white, fill opacity=1.0, solid},
                opacity=1.0,
            ] table [row sep=\\] {
                    x	y\\
                    2	27.41\\
                    4	27.38\\
                    6	27.52\\
                    8	27.54\\
                    10	27.57\\
                    12	27.51\\
                };
            \addplot+ [mark=*, draw=black, thick,
                mark size=1.25pt,
                mark options={fill=white, fill opacity=1.0, solid},
                opacity=1.0,
            ] table [row sep=\\] {
                    x	y\\
                    2	31.56\\
                    4	35.21\\
                    6	37.78\\
                    8	38.39\\
                    10	39.14\\
                    12	39.32\\
                };
            \addplot+ [mark=square*, draw=figure_red, thick,
                dashed,
                mark size=1.25pt,
                mark options={fill=white, fill opacity=1.0, solid},
                opacity=1.0,
            ] table [row sep=\\] {
                    x	y\\
                    2	36.59\\
                    4	40.36\\
                    6	40.25\\
                    8	40.16\\
                    10	40.02\\
                    12	39.82\\
                };
            \addplot+ [mark=*, draw=figure_blue, thick,
                mark size=1.25pt,
                mark options={fill=figure_blue, fill opacity=1.0, solid},
                opacity=1.0,
            ] table [row sep=\\] {
                    x	y\\
                    2	51.84\\
                    4	55.39\\
                    6	55.93\\
                    8	56.05\\
                    10	56.08\\
                    12	55.95\\
                };
        \end{groupplot}
        \node[anchor=north, legend] at (group c1r1.north) {\textbf{CSCD-NS} \textit{dev}};
        \node[anchor=north, legend] at (group c1r2.north) {\textbf{C2EC} \textit{dev}};
    \end{tikzpicture}
    \caption{
        Results of different beam sizes.
    }
    \label{fig:beam_size}
\end{figure}

Figure~\ref{fig:beam_size} shows the performance of our method with varying beam sizes on the CSCD-NS and C2EC datasets.
The results indicate that our method performs well even with a small beam size.
In particular, a beam size of 2 is sufficient to surpass \texttt{ICL}, \texttt{ICL-RR}, and \texttt{TfPf} on both datasets.
Interestingly, as the beam size increases, the performance of \texttt{ICL} remains almost unchanged, while the performance of \texttt{ICL-RR} steadily improves.
This suggests that while \texttt{ICL} can find a set of good candidates, it struggles to rank them properly.

\subsection{Impact of Two Rewards}
\label{app:ablation:two_reward}
\begin{table}[tb!]
    \centering
    \setlength{\tabcolsep}{4pt}%
    \scalebox{0.9}{
        \begin{NiceTabular}{lccc;ccc}
            \toprule
            \rowcolor[gray]{1.0}
            \Block[l]{2-1}{\textbf{System}} & \Block[c]{1-3}{\textbf{CSCD-NS} \textit{dev}} &            &                & \Block[c]{1-3}{\textbf{C2EC} \textit{dev}} &            &                \\
                                            & \textbf{P}                                    & \textbf{R} & \textbf{F$_1$} & \textbf{P}                                 & \textbf{R} & \textbf{F$_1$} \\
            \midrule

            \texttt{OUR}                    & 67.70                                         & 73.69      & 70.57          & 65.14                                      & 49.18      & 56.05          \\
            \hdashedline
            \ \ \texttt{-FR}                & 62.28                                         & 76.16      & 68.52          & 60.34                                      & 54.74      & 57.40          \\
            \ \ \texttt{-LR}                & 71.25                                         & 68.13      & 69.66          & 64.51                                      & 45.36      & 53.26          \\
            \bottomrule
        \end{NiceTabular}
    }
    \caption{
        Ablation study on the impact of two rewards.
    }
    \label{tab:ablation:two_reward}
\end{table}

In Equation \ref{eq:final_score}, we adopt both the faithfulness reward and the length reward from \texttt{TfPf} into the final scoring function.
This section investigates the effectiveness of these two rewards.
The ablation results are shown in Table~\ref{tab:ablation:two_reward}.

When the faithfulness reward is removed, our method shows an increase in recall, while the precision is reduced.
Conversely, removing the length reward results in better precision, but this comes with a decline in recall.

The faithfulness reward mainly improves precision, while the length reward mainly improves recall.
Together, they complement each other, leading to better overall performance.

\subsection{Adjusting the Precision-Recall Trade-off}
\label{app:precision_recall_trade_off}
\begin{figure}[t]
    \centering%
    \subfloat[\textbf{CSCD-NS} \textit{dev}]%
    {%
        \begin{tikzpicture}[trim axis left]
            \begin{axis}[
                    width=1.0\linewidth,
                    height=0.66\linewidth,
                    font=\scriptsize,
                    xlabel={Precision},
                    ylabel={Recall},
                    xmin=51, xmax=85,
                    ymin=51, ymax=81,
                    legend pos=south west,
                ]
                \addplot[
                    figure_blue,
                    thick,
                    mark=*,
                    mark size=2pt,
                    mark options={fill=figure_blue},
                ] coordinates {
                        (70.72,62.61) [0.50]
                        (71.82,65.07)
                        (72.19,66.68)
                        (70.92,69.23) [1.25]
                        (67.70,73.69) [1.50]
                        (60.47,77.10) [1.75]
                        (52.77,78.43) [2.00]
                    };

                \node[inner sep=1pt, font=\scriptsize, text=figure_blue, anchor=east]
                at (axis cs:70.00,62.61) {0.50};
                \node[inner sep=1pt, font=\scriptsize, text=figure_blue, anchor=east]
                at (axis cs:71.7,66.68) {1.00};
                \node[inner sep=1pt, font=\scriptsize, text=figure_blue, anchor=north east]
                at (axis cs:67.50,72.90) {1.50};
                \node[inner sep=1pt, font=\scriptsize, text=figure_blue, anchor=south]
                at (axis cs:60.47,77.75) {1.75};
                \node[inner sep=1pt, font=\scriptsize, text=figure_blue, anchor=south]
                at (axis cs:52.77,79.10) {2.00};

                \addplot[
                    figure_red,
                    thick,
                    mark=square*,
                    mark size=2pt,
                    mark options={draw=figure_red,fill=white},
                    point meta=explicit symbolic,
                    nodes near coords,
                    nodes near coords align={vertical},
                ] coordinates {
                        (37.60,75.84)
                        (56.51,75.92)
                        (67.70,73.69) [1.00]
                        (74.31,69.30) [1.25]
                        (78.69,64.06)
                        (81.78,58.34)
                        (84.28,52.27)
                    };

                \node[inner sep=1pt, font=\scriptsize, text=figure_red, anchor=north]
                at (axis cs:56.51,75.00) {0.75};
                \node[inner sep=1pt, font=\scriptsize, text=figure_red, anchor=south]
                at (axis cs:79.25,64.85) {1.50};
                \node[inner sep=1pt, font=\scriptsize, text=figure_red, anchor=south]
                at (axis cs:82.85,58.90) {1.75};
                \node[inner sep=1pt, font=\scriptsize, text=figure_red, anchor=east]
                at (axis cs:83.75,52.27) {2.00};

                \draw[draw=none, name path=auxline] (axis cs:51,51) -- (axis cs:85,81);

                \pgfplotsinvokeforeach{55.0,60.0,65.0,70.0,75.0,80.0,85.0} {
                    \pgfmathtruncatemacro{\curveid}{#1*10}

                    \addplot[gray, opacity=0.5, dashed, domain=51:85,
                        name path=curve\curveid] {#1 * x / (2 * x - #1)};

                    \path[name intersections={
                                of=auxline and curve\curveid,
                                name=inter\curveid,  %
                                total=\ttt
                            }];

                    \begin{scope}[on background layer]
                        \node[fill=white, inner sep=1pt, font=\tiny]
                        at (inter\curveid-1) {\pgfmathprintnumber{#1}};
                    \end{scope}
                }
            \end{axis}
        \end{tikzpicture}
    }%
    \\[5pt]
    \centering%
    \subfloat[\textbf{C2EC} \textit{dev}]
    {
        \begin{tikzpicture}[trim axis left]
            \begin{axis}[
                    width=1.0\linewidth,
                    height=0.91\linewidth,
                    font=\scriptsize,
                    xlabel={Precision},
                    ylabel={Recall},
                    xmin=49, xmax=76,
                    ymin=34, ymax=60,
                    legend pos=south west,
                ]
                \addplot[
                    figure_blue,
                    thick,
                    mark=*,
                    mark size=2pt,
                    mark options={fill=figure_blue},
                ] coordinates {
                        (70.53,39.89) [0.50]
                        (70.17,40.71)
                        (69.15,42.26) [1.00]
                        (67.46,44.17) [1.25]
                        (65.14,49.18) [1.50]
                        (58.85,56.65) [1.75]
                        (51.87,59.29) [2.00]
                    };

                \node[inner sep=1pt, font=\scriptsize, text=figure_blue, anchor=east]
                at (axis cs:70.00,39.89) {0.50};
                \node[inner sep=1pt, font=\scriptsize, text=figure_blue, anchor=east]
                at (axis cs:68.75,42.26) {1.00};
                \node[inner sep=1pt, font=\scriptsize, text=figure_blue, anchor=east]
                at (axis cs:67.00,44.17) {1.25};
                \node[fill=white, inner sep=1pt, font=\scriptsize, text=figure_blue, anchor=north east]
                at (axis cs:64.75,48.85) {1.50};
                \node[inner sep=1pt, font=\scriptsize, text=figure_blue, anchor=south]
                at (axis cs:58.85,57.25) {1.75};
                \node[inner sep=1pt, font=\scriptsize, text=figure_blue, anchor=north]
                at (axis cs:51.87,58.95) {2.00};

                \addplot[
                    figure_red,
                    thick,
                    mark=square*,
                    mark size=2pt,
                    mark options={draw=figure_red,fill=white},
                ] coordinates {
                        (39.72,56.83) [0.50]
                        (56.31,54.83) [0.75]
                        (65.14,49.18) [1.00]
                        (69.83,44.90) [1.25]
                        (72.01,40.53) [1.50]
                        (75.05,38.07) [1.75]
                        (75.54,35.16) [2.00]
                    };

                \node[inner sep=1pt, font=\scriptsize, text=figure_red, anchor=north]
                at (axis cs:56.31,54.40) {0.75};
                \node[inner sep=1pt, font=\scriptsize, text=figure_red, anchor=south west]
                at (axis cs:65.40,49.55) {1.00};
                \node[inner sep=1pt, font=\scriptsize, text=figure_red, anchor=south west]
                at (axis cs:70.10,44.75) {1.25};
                \node[inner sep=1pt, font=\scriptsize, text=figure_red, anchor=west]
                at (axis cs:72.45,40.65) {1.50};
                \node[inner sep=1pt, font=\scriptsize, text=figure_red, anchor=east]
                at (axis cs:74.50,38.07) {1.75};
                \node[inner sep=1pt, font=\scriptsize, text=figure_red, anchor=east]
                at (axis cs:75.00,35.16) {2.00};

                \draw[draw=none, name path=auxline] (axis cs:49,34) -- (axis cs:76,60);

                \pgfplotsinvokeforeach{45.0,50.0,55.0,60.0,65.0} {
                    \pgfmathtruncatemacro{\curveid}{#1*10}

                    \addplot[gray, opacity=0.5, dashed, domain=49:76,
                        name path=curve\curveid] {#1 * x / (2 * x - #1)};

                    \path[name intersections={
                                of=auxline and curve\curveid,
                                name=inter\curveid,  %
                                total=\ttt
                            }];

                    \begin{scope}[on background layer]
                        \node[fill=white, inner sep=1pt, font=\tiny]
                        at (inter\curveid-1) {\pgfmathprintnumber{#1}};
                    \end{scope}
                    \legend{
                        Temperature,
                        $\gamma$,
                    }
                }
            \end{axis}
        \end{tikzpicture}
    }
    \caption{
        Precision-Recall trade-off curve with F1 score iso-contours.
        The curve shows the relationship between precision and recall, while the dashed lines represent constant F1 score values.
        The values near the nodes are the temperature values.
        In main experiments, we set the temperature to 1.50 and $\gamma$ to 1.00.
    }
    \label{fig:pr_curve}
\end{figure}

In real-world applications, depending on user needs, higher recall may be preferred over precision, or vice versa.
For example, newspaper editors, who can verify the correctness of model-generated corrections, might prefer higher recall to identify as many potential errors as possible.

Our method offers two ways to adjust the precision-recall trade-off:
\begin{inparaenum}[1)]
    \item Adjusting the temperature of the prompt-based LLM.
    \item Introducing a new parameter, $\gamma$, as a coefficient for the distance metric term, modifying Equation~\ref{eq:gcsc_score} as follows:
\end{inparaenum}
\begin{equation}
    \begin{aligned}
        s(\substring{x}, \substring{y}) =\  & {\log p_{\mathtt{LLM}}(\substring{y}\mid \mathtt{Prompt}(\substring{x}))}                                 \\
                                            & + \log p_{\mathtt{LLM}}(\substring{y}) - \gamma\ \mathtt{Dist}_{\mathtt{L}}(\substring{x}, \substring{y})
    \end{aligned}
\end{equation}

Figure~\ref{fig:pr_curve} illustrates the precision-recall trade-off curve with F1 score iso-contours.

As observed in Figure~\ref{fig:pr_curve}, for temperature values between 1.0 and 2.0, increasing the temperature leads to higher recall but lower precision.
On the CSCD-NS dataset, using a temperature value below 1.0 results in a decrease in both precision and recall.

The $\gamma$ parameter, which controls the influence of the distance metric, has an opposite effect on the precision-recall trade-off.
A higher $\gamma$ value results in higher precision but lower recall.

The figure indicates that the $\gamma$ parameter is more effective for increasing precision, whereas the temperature parameter is more effective for increasing recall.

\begin{table}[tb!]
    \centering
    \scalebox{0.9}{
        \begin{NiceTabular}{lccc}
            \toprule
            \rowcolor[gray]{1.0}
            \textbf{System}                               & \textit{per} \textbf{Sent.} & \textit{per} \textbf{Char.} & \textbf{Mem} \\
            \midrule
            \Block[c]{1-4}{\textbf{Length:} $< 32$}       &                             &                             &              \\
            \midrule
            \texttt{ICL (Greedy)}                         & \wz703.3                    & 30.5                        & 15650        \\
            \texttt{ICL}                                  & 1379.2                      & 59.8                        & 20025        \\
            \texttt{TfPf}                                 & 1118.3                      & 48.5                        & 16404        \\
            \hdashedline
            \texttt{Our}                                  & 1944.5                      & 84.3                        & 16670        \\
            \midrule
            \Block[c]{1-4}{\textbf{Length:} $32 \sim 64$} &                             &                             &              \\
            \midrule
            \texttt{ICL (Greedy)}                         & 1322.4                      & 28.3                        & 15677        \\
            \texttt{ICL}                                  & 1957.9                      & 41.9                        & 20234        \\
            \texttt{TfPf}                                 & 1951.4                      & 41.8                        & 16411        \\
            \hdashedline
            \texttt{Our}                                  & 3715.1                      & 79.6                        & 16792        \\
            \midrule
            \Block[c]{1-4}{\textbf{Length:} $> 64$}       &                             &                             &              \\
            \midrule
            \texttt{ICL (Greedy)}                         & 2182.9                      & 26.7                        & 15684        \\
            \texttt{ICL}                                  & 2894.8                      & 35.4                        & 20297        \\
            \texttt{TfPf}                                 & 3222.5                      & 39.4                        & 16421        \\
            \hdashedline
            \texttt{Our}                                  & 6155.7                      & 75.3                        & 16956        \\
            \bottomrule
        \end{NiceTabular}
    }
    \caption{
        Runtime and memory usage comparison between baseline and our method.
        The unit of time is ms, and the unit of memory is MB.
    }
    \label{tab:analysis:runtime}
\end{table}

\subsection{Run-time Analysis}
\label{app:runtime}
We randomly sampled 50 sentences of varying lengths from the development set of CSCD-NS and C2EC to evaluate the running time of our method on \texttt{Qwen2.5\,7B} compared to the baselines.
The experiment was conducted on a single NVIDIA A100 40GB GPU with the Intel Xeon Gold 6248R (3.00GHz) CPU.
The batch size was set to 1 during evaluation.
The results, as shown in Table~\ref{tab:analysis:runtime}, indicate that our approach is approximately twice slower than \texttt{TfPf}.
This increased time is due to the two forward passes of the large language model at each step to obtain the final score.
However, since these two forward passes are independent, the process can be accelerated by parallelizing them if more GPUs are available.

\begin{figure*}[p!]
    \captionsetup[subfigure]{skip=0pt}
    \centering%
    \begin{promptbox}{\texttt{CSC In-Context Prompt for Baseline}}{black}{white}
        \footnotesize{
            \textbf{\texttt{System Prompt:}}\\
            你是一个优秀的中文纠错模型，中文纠错模型即更正用户输入句子中的错误。\return

            \textbf{\texttt{User Prompt:}}\\
            你需要识别并纠正用户输入的句子中可能的错别字并输出正确的句子，在纠正错别字的同时尽可能减少对原句子的改动(不新增、删除和修改标点符号)。只输出没有错误的句子，不要添加任何其他解释或说明。如果句子没有错误，就直接输出和输入相同的句子。\return\return

            \jsonkey{<Example>}\return\\
            输入：\jsonkey{\{INPUT\_EXAMPLE\_1:\  $\substring{x}_1$\}}\return\\
            输出：\jsonkey{\{OUTPUT\_EXAMPLE\_1: $\substring{y}_1$\}}\return\return\\
            输入：\jsonkey{\{INPUT\_EXAMPLE\_2:\  $\substring{x}_2$\}}\return\\
            输出：\jsonkey{\{OUTPUT\_EXAMPLE\_2: $\substring{y}_2$\}}\return\return\\
            \jsonkey{</Example>}\return\return\\

            输入：\jsonkey{\{INPUT: $\substring{x}$\}}\return
            输出：
        }
        \tcblower
        \scriptsize
        \textbf{\texttt{System Prompt:}}\\
        You are an excellent Chinese error correction model, which means you correct errors in the sentences entered by the user.\return

        \textbf{\texttt{User Prompt:}}\\
        You need to identify and correct possible misspelled characters in the user's input sentence, and output the correct sentence. While making corrections, try to minimize changes to the original sentence (without adding, deleting, or modifying punctuation). Only output the corrected sentence; do not add any further explanations or notes. If the sentence is error-free, output the exact same sentence as the input.\return

        \jsonkey{<Example>}\return\\
        \rlap{Input:}\phantom{Output:} \jsonkey{\{INPUT\_EXAMPLE\_1:  $\substring{x}_1$\}}\return\\
        Output: \jsonkey{\{OUTPUT\_EXAMPLE\_1: $\substring{y}_1$\}}\return\return\\
        \rlap{Input:}\phantom{Output:} \jsonkey{\{INPUT\_EXAMPLE\_2:  $\substring{x}_2$\}}\return\\
        Output: \jsonkey{\{OUTPUT\_EXAMPLE\_2: $\substring{y}_2$\}}\return\return\\
        \jsonkey{</Example>}\return\return\\\\
        \rlap{Input:}\phantom{Output:} \jsonkey{\{INPUT: $\substring{x}$\}}\return\\
        Output:

    \end{promptbox}
    \begin{promptbox}{\texttt{C2EC In-Context Prompt for Baseline}}{black}{white}
        \footnotesize{
            \textbf{\texttt{System Prompt:}}\\
            你是一个优秀的中文纠错模型，中文纠错模型即更正用户输入句子中的错误。\return

            \textbf{\texttt{User Prompt:}}\\
            你需要识别并纠正用户输入的句子中可能的\correct{错别字、多字、漏字}并输出正确的句子，在修改的同时尽可能减少对原句子的改动(不新增、删除和修改标点符号)。只输出没有错误的句子，不要添加任何其他解释或说明。如果句子没有错误，就直接输出和输入相同的句子。\return\return

            \jsonkey{<Example>}\return\\
            输入：\jsonkey{\{INPUT\_EXAMPLE\_1:\  $\substring{x}_1$\}}\return\\
            输出：\jsonkey{\{OUTPUT\_EXAMPLE\_1: $\substring{y}_1$\}}\return\return\\
            输入：\jsonkey{\{INPUT\_EXAMPLE\_2:\  $\substring{x}_2$\}}\return\\
            输出：\jsonkey{\{OUTPUT\_EXAMPLE\_2: $\substring{y}_2$\}}\return\return\\
            \jsonkey{</Example>}\return\return\\

            输入：\jsonkey{\{INPUT: $\substring{x}$\}}\return
            输出：
        }
        \tcblower
        \scriptsize
        \textbf{\texttt{System Prompt:}}\\
        You are an excellent Chinese error correction model, which means you correct errors in the sentences entered by the user.\return

        \textbf{\texttt{User Prompt:}}\\
        You need to identify and correct possible misspellings, redundant and missing characters in the user's input sentence, and output the correct sentence. While making corrections, try to minimize changes to the original sentence (without adding, deleting, or modifying punctuation). Only output the corrected sentence; do not add any further explanations or notes. If the sentence is error-free, output the exact same sentence as the input.\return

        \jsonkey{<Example>}\return\\
        \rlap{Input:}\phantom{Output:} \jsonkey{\{INPUT\_EXAMPLE\_1:  $\substring{x}_1$\}}\return\\
        Output: \jsonkey{\{OUTPUT\_EXAMPLE\_1: $\substring{y}_1$\}}\return\return\\
        \rlap{Input:}\phantom{Output:} \jsonkey{\{INPUT\_EXAMPLE\_2:  $\substring{x}_2$\}}\return\\
        Output: \jsonkey{\{OUTPUT\_EXAMPLE\_2: $\substring{y}_2$\}}\return\return\\
        \jsonkey{</Example>}\return\return\\\\
        \rlap{Input:}\phantom{Output:} \jsonkey{\{INPUT: $\substring{x}$\}}\return\\
        Output:

    \end{promptbox}
    \caption{
        In-context learning prompts for baseline models.
    }
    \label{fig:icl_detail_prompt}
\end{figure*}

\end{CJK}
\end{document}